%% file: arxiv.tex
\documentclass[11pt]{article}

\usepackage[preprint]{acl}

\usepackage{times}
\usepackage{latexsym}

\usepackage[T1]{fontenc}

\usepackage[utf8]{inputenc}

\usepackage{microtype}

\usepackage{inconsolata}

\usepackage{graphicx}
\usepackage{amsmath} 
\usepackage{amsfonts}
\usepackage{enumitem}
\usepackage{soul}
\usepackage{booktabs,tabularx}
\usepackage{comment}
\usepackage{mdframed}
\usepackage{subcaption}  
\usepackage{float}       
\usepackage{hyperref}
\usepackage{color}

\newcommand{\icon}[2]{%
  \raisebox{-0.3em}{\includegraphics[height=1.3em]{figs/icons/#1.png}}%
}

\usepackage{tcolorbox}       
\usepackage{xcolor}          
\definecolor{myBlue}{RGB}{0, 102, 204}  

\usepackage[table]{xcolor}  

\title{What Matters to an LLM? Behavioral and Computational Evidences from Summarization}

\author{Yongxin Zhou, Changshun Wu, Philippe Mulhem, Didier Schwab, Maxime Peyrard \\
        Univ. Grenoble Alpes, CNRS, Inria, Grenoble INP, LIG, 38000, Grenoble, France \\
        \texttt{\{firstname.lastname\}@univ-grenoble-alpes.fr}
       }
       
\begin{document}
\maketitle
\begin{abstract}
Large Language Models (LLMs) are now state-of-the-art at summarization, yet the internal notion of importance that drives their information selections remains hidden. We propose to investigate this by combining behavioral and computational analyses. Behaviorally, we generate a series of length-controlled summaries for each document and derive empirical importance distributions based on how often each information unit is selected. These reveal that LLMs converge on consistent importance patterns, sharply different from pre-LLM baselines, and that LLMs cluster more by family than by size. Computationally, we identify that certain attention heads align well with empirical importance distributions, and that middle-to-late layers are strongly predictive of importance. Together, these results provide initial insights into \emph{what} LLMs prioritize in summarization and \emph{how} this priority is internally represented, opening a path toward interpreting and ultimately controlling information selection in these models.\footnote{Our code and data are available at \url{https://github.com/yongxin2020/llm-importance-summarization}.}
\end{abstract}

\section{Introduction}
Large Language Models (LLMs) are increasingly entrusted with the management of information: they filter, select, and summarize our textural information. This shift raises a fundamental question: \textit{what do LLMs consider important?} Summarization offers a natural entry point to this question, since the core challenge of summarization is the identification and selection of \textit{important information}.

Before the rise of LLMs, summarization systems were designed around simple, surface-level heuristics that correlated with importance as reflected in benchmark datasets. For example, centrality measures~\cite{zopf-etal-2016-beyond}, lexical overlap~\cite{luhn1958automatic}, and frequency of mention~\cite{edmundson1969new,nenkova2005impact} were strong predictors of which content would be selected. Remarkably, the crude baseline of taking the first few sentences of a news article remained one of the hardest to beat for years~\cite{see-etal-2017-get,xing-etal-2021-demoting}.
LLMs radically changed this landscape. Today, summarization is largely accomplished by prompting an LLM, yielding fluent, flexible, and adaptable summaries with little or no task-specific engineering~\cite{zhang-etal-2024-benchmarking}. Yet this success comes at the cost of transparency: unlike extractive methods that explicitly score information units~\citep{10.5555/1763653.1763720,gillick2008icsi,li-etal-2013-using,peyrard-eckle-kohler-2016-optimizing}, LLM-based summarization provides no clear account of the internal notion of importance driving its selections. Understanding this hidden notion of importance is both scientifically and practically critical, as it speaks to how LLMs structure and prioritize knowledge~\citep{carter2019activation,cammarata2020thread,NEURIPS2022_6f1d43d5,geva-etal-2023-dissecting,monea-etal-2024-glitch} in ways that increasingly shape human access to information. 

\begin{figure*}[!htb]
\centering
    \includegraphics[trim=0 4.5cm 0 0.2cm, clip, width=\textwidth]{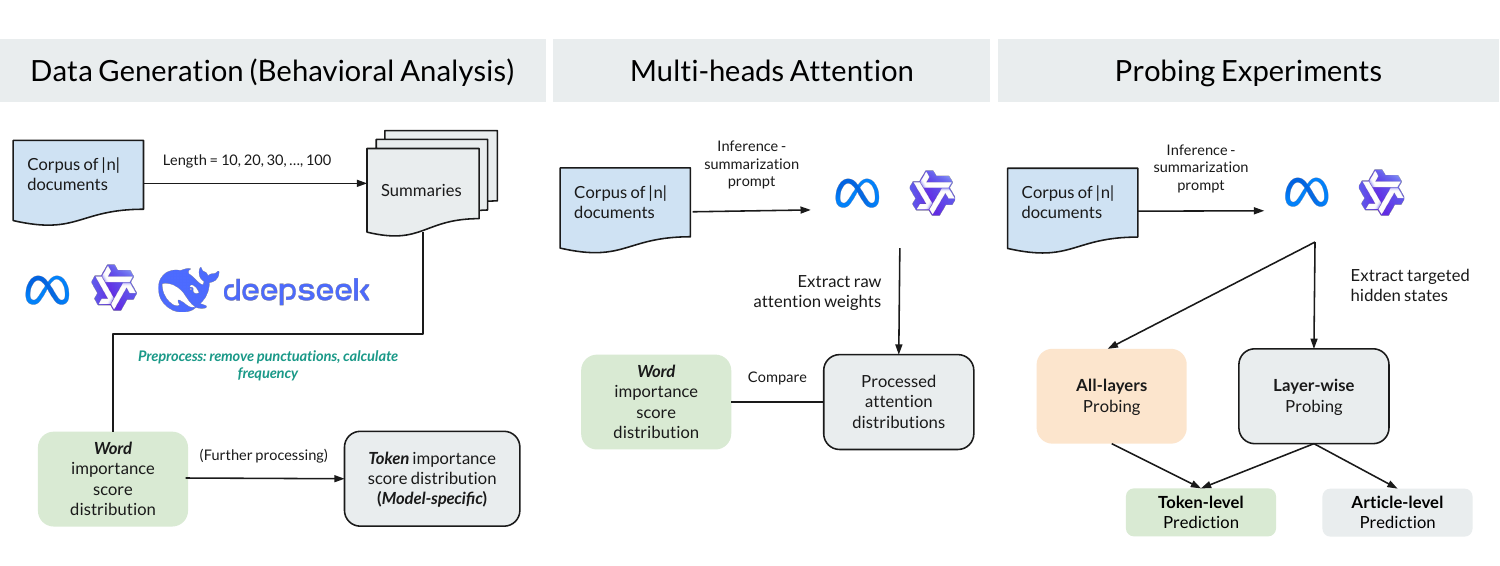}
    \caption{Analytical framework for modeling \emph{information importance}. (1) \textbf{Behavioral Analysis:} We generate length-variant summaries with LLMs across three datasets (CNN/DailyMail, SAMSum, DECODA-French).
    The \emph{importance distribution} \( I_M(D) \) is derived as summary persistence. (2) \textbf{Attention Analysis:} Raw attention weights are aggregated and normalized to obtain token-level distributions.
    (3) \textbf{Probing:} Hidden states are used to train probes in three scenarios (S1: Layer-wise/Token, S2: All-layers/Token, S3: Layer-wise/Article) to predict \( I_M(D) \).}
    \label{fig:overview}
\end{figure*}

In this work, we take a two-pronged interpretability approach to this question visualized in Figure~\ref{fig:overview}. With \textbf{behavioral analysis}, we first study model outputs themselves. For any given input document, we generate 10 summaries of varying lengths, then track how frequently each information unit (e.g., token) is selected across summaries. This yields an empirical importance distribution: an estimate of what the model consistently prioritizes when forced to compress information at different lengths.
With the \textbf{computational analysis}, we then turn inward to the model’s computation, asking whether and where this empirical importance is represented. Attention mechanisms, central to the Transformer architecture, are themselves distributions over tokens and can be interpreted as mechanisms of importance weighting~\citep{kobayashi-etal-2020-attention}. We therefore test whether specific attention heads align with the behavioral importance distributions. Beyond attention, we probe layer activations: using a low-capacity linear probe specifically designed to output importance scores for all tokens simultaneously. This setup mitigates common risks of overfitting in probing studies~\citep{ravichander-etal-2021-probing,kumar2022probing,exp_unders}. Furthermore, to avoid false discoveries we include a ``dead salmon'' probe baseline by comparing probe predictions against a probe trained on a randomized version of the models~\citep{méloux2025deadsalmonsaiinterpretability}.

\textbf{Contributions.}
We release a dataset of 274,330 summaries generated by 7 LLMs, preprocessed to obtain the empirical importance distributions. 
Our behavioral analysis shows that LLMs exhibit broadly similar summarization behavior, sharply diverging from pre-LLM baselines. 
Similarity is influenced more by model family than by scale.
Our computational analysis demonstrates that certain attention heads alone capture substantial aspects of the importance distribution, and that middle-to-late layers of the network are highly predictive of behavioral importance. 
This work initiates a research direction to better understand, and ultimately control, the latent notion of information importance encoded within the computational structure of LLMs.

\section{Related Work}
\subsection{Information Importance in Summarization}
\label{subsec:importance_in_summarization}

The core task of summarization is to select and condense salient information from a source document, a process that fundamentally requires learning a latent representation of \emph{information importance}, an abstract and context-dependent concept \citep{narayan-etal-2018-dont, kedzie-etal-2018-content, peyrard-gurevych-2018-objective}.
Historically, \emph{importance} was operationalized through surface-level statistical features for extractive methods, such as term frequency and sentence position \citep{luhn1958automatic}, or measured post-hoc by $n$-gram overlap metrics like ROUGE against a reference \citep{lin-2004-rouge}. While these approaches facilitated practical solutions, they captured signals that merely correlate with human intuition. For instance, structural features like centrality and repetitions remain common proxies \citep{kedzie-etal-2018-content}, but their weaknesses are exposed by simple adversarial attacks \citep{zopf-etal-2016-beyond}. 
Theoretical efforts provided a more formal foundation. Early work treated \emph{importance} as a latent variable optimized indirectly via system performance \citep{Nenkova2012}. A deeper formalization grounds it in information theory, representing texts as probability distributions over semantic units \citep{Bao2011TowardsAT}. This view, compatible with distributional embeddings, allows information-theoretic tools to operate at a semantic level \citep{Carnap1952AnOO, IS4SI-2017-04000}. \citet{peyrard-2019-simple} crystallizes this framework by formally defining \emph{importance} through the unified concepts of redundancy, relevance, and informativeness.

Empirical studies investigate how models operationalize this latent construct. For instance, behavioral probes using length-controlled summarization show that LLMs develop a nuanced, hierarchical understanding of salience, though this internal representation shows weak correlation with human judgment and is not directly accessible via introspection \citep{trienes-etal-2025-behavioral}. The effective definition of \emph{importance} is not universal but is contingent on the summarization task and its communicative goals \citep{zhou-etal-2025-gpt}. It is therefore shaped by conversational dynamics and action-oriented objectives in dialogues \citep{zhou-etal-2024-psentscore, ghebriout-etal-2025-quartz} and by event centrality in news \citep{li-etal-2016-abstractive, zhang-etal-2024-benchmarking}.

A key open question remains how this abstract concept is mechanistically \emph{encoded} within models. Our work addresses this gap through a multi-faceted analysis: (1) a \textbf{behavioral} study of summary outputs, (2) an examination of \textbf{attention} mechanisms, and (3) \textbf{probing} of hidden state representations to trace how a model's own output-based importance signal is constructed across layers.

\subsection{Behaviors and Biases in Summarization}

Research on summarization behaviors and biases investigates what content models prioritize and how faithfully they reproduce it. A well-documented behavioral bias is \textbf{positional bias}, where models under-attend to middle content, creating a ``U-shaped'' trend in faithfulness for long-form summarization \citep{wan-etal-2025-positional} that also impacts conversational summarization by affecting how models handle information based on its location in a dialogue \citep{sun-etal-2025-posum}.
 
Beyond content selection, studies analyze biases in how content is generated. This includes analyzing \textbf{social biases} in summaries, where controlled evaluations isolate model bias from source bias to measure issues like demographic or gender skews in generated content \citep{steen-markert-2024-bias}. It also includes the analysis of \textbf{faithfulness and factuality} \citep{maynez-etal-2020-faithfulness, goyal-durrett-2021-annotating}, a core challenge where LLMs exhibit distinct error patterns, such as generating plausible ``circumstantial'' inferences in dialogues unsupported by direct evidence \citep{ramprasad-etal-2024-analyzing}.

While existing research documents \emph{what} models output, including systematic biases like positional effects, the internal computational origins of these behaviors remain unclear. Our work investigates these internal mechanisms alongside a behavioral analysis of summarization outputs.

\subsection{Interpretability and Probing for Analysis}

The multi-head attention mechanism, introduced by \citet{NIPS2017_3f5ee243}, enables transformers to attend to information from different representation subspaces. Analyses indicate that attention heads often specialize in specific linguistic phenomena \citep{voita-etal-2019-analyzing}, and their distributions have consequently been interpreted as a form of token importance weighting and used as a window into model behavior and decision-making \citep{kobayashi-etal-2020-attention, li-etal-2022-human}.

A parallel research direction seeks to understand the rich information encoded within LLMs' hidden representations, which frequently surpasses what is expressed in their explicit outputs \citep{burns2023discovering}. For instance, probing classifiers can decode latent knowledge such as pre-encoded plans for future response \citep{dong2025emergent}, internal signals of factual inaccuracies \citep{orgad2025llms}, or layer-wise capabilities for simulating personality traits, which can subsequently be leveraged to edit the personality expressed by LLMs during inference \citep{ju2025probing}.

While existing work has investigated attention specialization and applied probing to tasks like syntax analysis or fact-checking, the mechanistic representation of \emph{summarization-specific} importance remains less explored. Our work bridges these areas by analyzing internal representations to trace how a model's own operational definition of \emph{importance} is computed and encoded throughout the transformer architecture.

\section{Methodology}
\label{sec:methodology}
Let \( M \) be a fixed transformer-based language model and let \( D = (u_1, \dots, u_n) \) denote a document decomposed into a set of atomic \emph{information units} (e.g., sentences, discourse units, or tokens in the rest of the paper).  
We introduce a model-dependent \emph{importance distribution} \( I_M(D) \), which captures the relative importance assigned by model \( M \) to the information units in \( D \).

Formally, \( I_M(D) \) is a probability distribution over the units of \( D \), where \( I_M(D)_j \) reflects the propensity of model \( M \) to include unit \( u_j \) when constrained to produce a summary of limited length.  This distribution is intended to encode the model’s implicit trade-offs and preferences when selecting a small subset of units to represent the document.

Because \( I_M(D) \) is not directly observable, we estimate it empirically via repeated conditional generation.  
Given a document \( D \), we prompt \( M \) to generate \( k \) summaries of varying target lengths, with \( k = 10 \) in our experiments.  Let \( S^{(\ell)} = \{s^{(\ell)}_1, \dots, s^{(\ell)}_k\} \) denote the set of summaries generated at length constraint \( \ell \).

For each information unit \( u_j \), we compute its empirical selection frequency
\begin{equation}
\label{eq:importance}
\hat{I}_M(D)_j = \frac{1}{k} \sum_{i=1}^{k} \mathbb{I}\big[u_j \in s_i\big],
\end{equation}
where \( \mathbb{I}[\cdot] \) is the indicator function and \( s_i \) ranges over all generated summaries across lengths.  Intuitively, units that appear consistently in shorter summaries and persist in longer ones receive higher importance scores. This principle is supported by Pyramid-based evaluation methods, which also count the frequency of information in human reference summaries as a proxy for importance~\citep{10.1145/1233912.1233913}.

Using the estimated importance distribution \( I_M(D) \), we pursue three complementary objectives:
(i) To analyze the behavioral patterns of summarization models by studying the properties of \( I_M(D) \) across datasets (Section~\ref{sec:behavioral});
(ii) To investigate the relationship between empirical importance \( I_M(D) \) and the model’s internal attention mechanisms (Section~\ref{sec:multi_head_attention});
(iii) To assess whether importance information is encoded in the model’s hidden state representations via probing methods (Section~\ref{sec:probing}).

\subsection{Datasets}
\label{subsubsec:datasets}

We conduct experiments on two core summarization datasets to derive and analyze the empirical importance distribution. Their contrasting characteristics allow us to evaluate whether patterns of importance encoding are consistent across different genres and summary styles.

\begin{itemize}[noitemsep,topsep=0pt,parsep=0pt,partopsep=0pt]
    \item \textbf{CNN/DailyMail} \citep{see-etal-2017-get}: An English news dataset containing over 300k journalist-written articles. We use version 3.0.0\footnote{\url{https://huggingface.co/datasets/ccdv/cnn_dailymail}}, originally designed for reading comprehension and question answering but widely adopted for extractive and abstractive summarization.
    \item \textbf{SAMSum} \citep{gliwa-etal-2019-samsum}: A dataset of $\sim$16k messenger-style English conversations with linguist-created summaries. It includes 14,732 training, 818 validation, and 819 test samples, providing a testbed for abstractive summarization of informal, interactive text.
\end{itemize}

\textbf{Multilingual Extension:} To assess the generalizability of our findings across languages, we additionally evaluate on the \textbf{DECODA} corpus \citep{favre-etal-2015-call}, a French call center dialogue dataset. The corresponding experiments and analyses are provided in Appendix~\ref{appendix:french_decoda} due to space constraints.

\subsection{Models}

We evaluate the following models: two open-weight model families and one commercial model. 

\begin{itemize}[noitemsep,topsep=0pt,parsep=0pt,partopsep=0pt]
    \item \icon{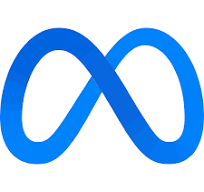}~~Llama-3.2-1B-Instruct, Llama-3.1-8B-Instruct
    \item \icon{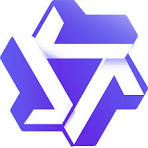}~~Qwen2.5-1.5B-Instruct, Qwen2.5-3B-Instruct, Qwen2.5-7B-Instruct, Qwen2.5-14B-Instruct
    \item \icon{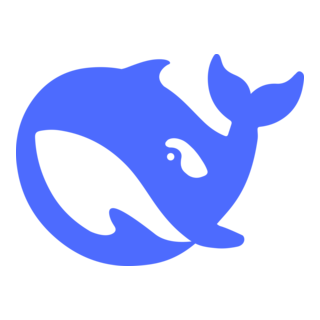}~~deepseek-chat\footnote{Experiments conducted in September 2025, the model points to non-thinking mode of DeepSeek-V3.1.}
\end{itemize}

The models and their corresponding links are detailed in Appendix \ref{appendix:model_specification}.

\subsection{Data Generation}

We generated data using 3,000 samples from the CNN/DailyMail test set, the full SAMSum test set (819 samples), and the full DECODA test set (100 samples). For each input, we generated $k=10$ summaries of varying lengths using length-variant prompts (see Table \ref{tab:prompt_data_generation}, where $N \in \{10, 20, 30, \dots, 100\}$). We then compute the empirical importance distribution for each model and each document using the formula described in Equation~\ref{eq:importance}.

\subsection{Metric Selection and Validation}

To compare importance distributions, several classes of metrics are applicable. These include (i) distributional measures such as Kullback--Leibler divergence, Jensen--Shannon divergence, and Wasserstein distance; (ii) list comparison measures such as Pearson and Spearman correlations; and (iii) information retrieval metrics that emphasize agreement among top-ranked units, including nDCG@\(k\). 
In the main experiments, we focus on two complementary metrics: Spearman correlation and nDCG@10. When a specific reference importance distribution is available (e.g., in attention-based and probing analyses), we report nDCG@10.  
In settings without a designated reference distribution, we use the symmetric Spearman correlation as a notion of similarity.
However, we evaluate a total of 14 comparison metrics, and Appendix~\ref{appendix:metrics} provides more details on our choice of Spearman and nDCG. Importantly, our findings are qualitatively robust to the choice of metric.  
Results obtained with alternative metrics are reported in the Appendix for behavioral analysis (Appendix~\ref{appendix:behavioral_analysis}), multi-head attention analysis (Appendix~\ref{appendix:multi_head_attention}), and probing experiments (Appendix~\ref{appendix:probing_all_scenarios}).

\section{Behavioral Analysis}
\label{sec:behavioral}

This section analyzes how the \textit{summary persistence} importance distribution \(I_M(D)\) varies across model families and scales. We quantify inter-model similarities and identify recurring statistical patterns in model outputs to characterize their shared notion of importance. Additional analyses of positional bias and entropy are provided in Appendix~\ref{appendix:positional_bias} and Appendix~\ref{appendix:model_entropy}, respectively.

\subsection{Summarization Baselines}
\label{subsec:baseline_approaches}

To compare LLM summarization behavior, we implement three pre-LLM baselines:

\textbf{Baseline 1: First-N-Words Frequency.} 
Simulates lead bias by calculating word frequency across ten document truncations (first 10, 20, ..., 100 words). This tests if LLMs prioritize early content beyond simple heuristics.

\textbf{Baseline 2: Token Frequency.}
Estimates importance using raw word counts normalized by the document's maximum word frequency, serving as a basic statistical baseline.

\textbf{Baseline 3: TextRank.}
Extracts and scores keywords using the TextRank algorithm~\citep{mihalcea-tarau-2004-textrank}, with scores normalized to the range [0, 1] via min-max normalization.

We also introduce \textbf{Human Frequency}, which assigns importance scores based on word presence in ground-truth summaries
\footnote{Note: This serves as a \textbf{proxy} for human-annotated importance derived from reference summaries, rather than direct human annotation of source word importance.}.

\subsection{Model Behavioral Similarity Analysis}
\label{subsec:model_similarity}

\begin{figure}[!htb]
  \centering
  
    \includegraphics[width=\linewidth]{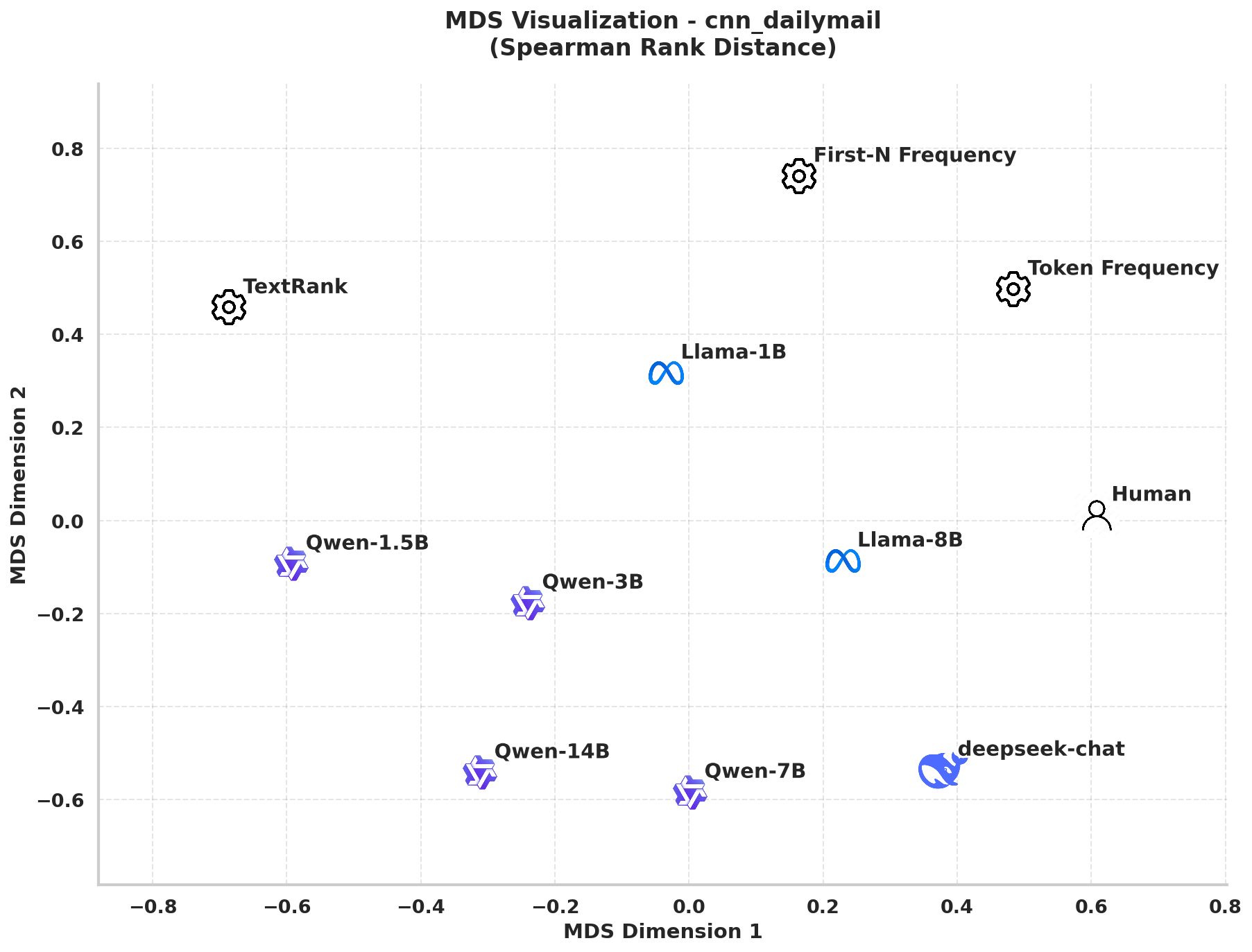}
    \vspace{0.1cm}
    \includegraphics[width=\linewidth]{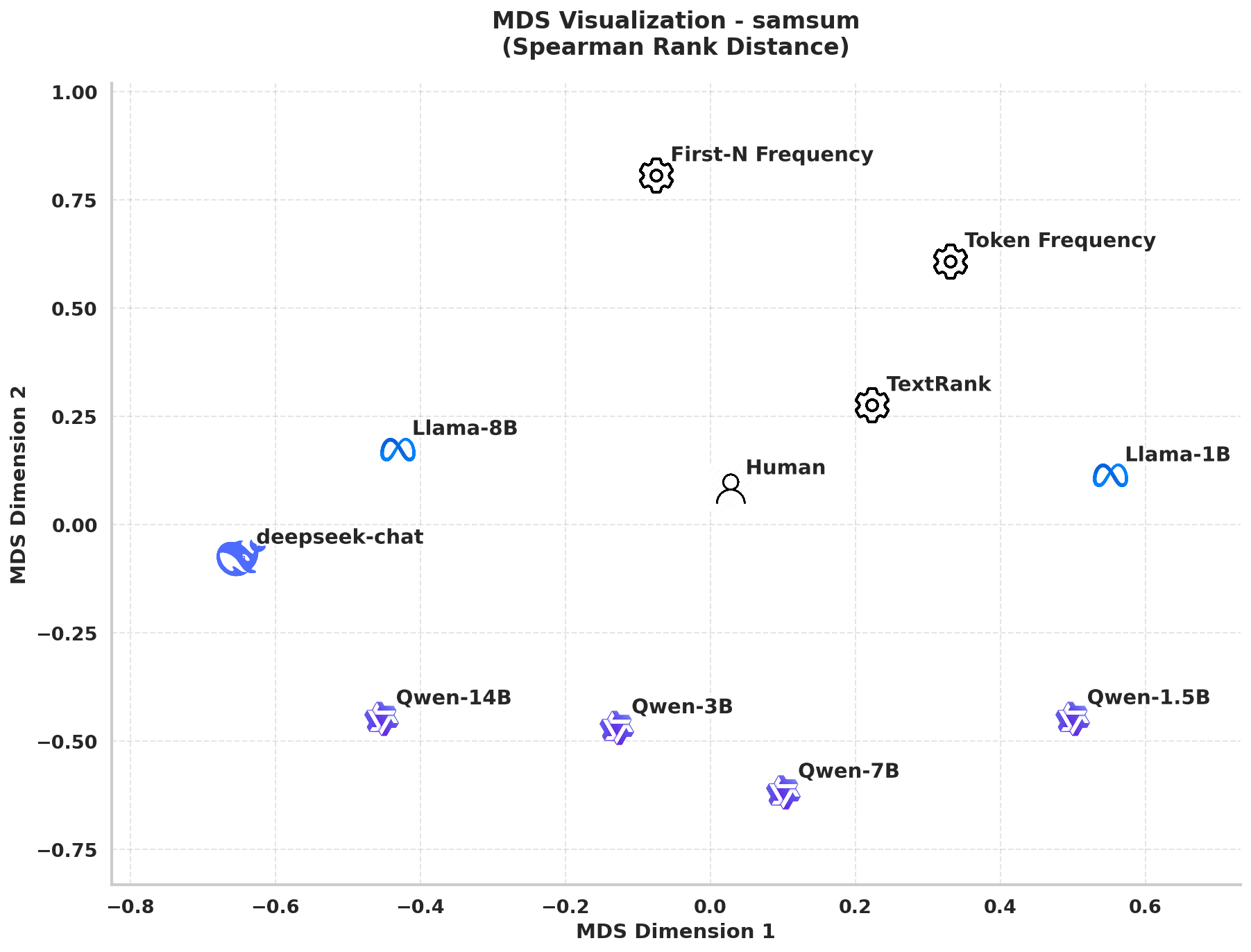}
  \caption{Pairwise model similarity based on Spearman rank correlation distance for importance distributions, visualized via two-dimensional Multidimensional Scaling (MDS). Results are shown for the CNN/DailyMail (top) and SAMSum (bottom) datasets.}
  \label{fig:model_similarity_spearman}
\end{figure}

To analyze similarity in importance distributions, we construct a union vocabulary from all words appearing in any model's summaries for each document. For each document, words present in one model's importance distribution but absent from another's were assigned an importance score of zero, ensuring valid probability distributions.

We calculate pairwise model \textit{dissimilarity} using Spearman rank correlation distance, defined as $d = 1 - \rho$, where $\rho$ is Spearman's correlation coefficient. For each document, we compute the Spearman distance between models' word importance rankings, then average across all common documents to obtain a single pairwise distance. 
These distances are visualized via Multidimensional Scaling (MDS) in Figure~\ref{fig:model_similarity_spearman}, which projects the dissimilarity values into two-dimensional Euclidean space where inter-model distances reflect their behavioral similarity. Our analysis reveals two key patterns in summarization behavior:
\begin{enumerate}[leftmargin=*, noitemsep, topsep=3pt]
    \item \textbf{Distinct LLM Clustering:} LLMs exhibit broadly similar behavior, forming a cluster distinct from pre‑LLM baselines. \textit{Human Frequency} occupies a central position between LLM and pre-LLM summarization models.
    \item \textbf{Architectural Bias:} Models tend to cluster by family (e.g., Llama vs. Qwen). This family-based clustering appears more pronounced on the CNN/DailyMail dataset than on SAMSum.
\end{enumerate}

The distributional similarity measured by NDCG@10 is provided in Figure~\ref{fig:model_similarity_ndcg} in the Appendix.

\subsection{Qualitative Insights}
\label{subsubsec:qualitative_insights}

The importance distribution \(I_M(D)\) across all models and datasets (Figure~\ref{fig:importance_score_grouped_bar}, Appendix) shows that most words receive low scores, with approximately 50\% of word-importance pairs assigned a score of 0.1. Only a small fraction of words are marked as highly important ($\geq$0.8).

In both datasets, \textbf{named entities and core concepts} (e.g., main actors, central events) consistently receive the highest importance scores. \textbf{Function words and general connectors} are uniformly assigned low importance. Supporting details typically fall within a medium importance range. A more detailed analysis of score distributions is provided in Appendix~\ref{appendix:importance_score_analysis}.

\section{Attention–Importance Alignment}
\label{sec:multi_head_attention}

We now examine internal computations to address the second goal: \textit{How closely do a model's multi-head attention patterns align with the output-based importance distribution \( I_M(D) \)?}

For each sample (300 from CNN/DailyMail, 819 from SAMSum), we extract the raw attention weights from the model's multi-head attention layers when processing the input prompt (before generating the summary). Then, we compute a token-level \textit{attention received} score by summing, for each token, the incoming attention from all other positions in the sequence.
For words composed of multiple subword tokens, we averaged the attention scores of their constituent tokens. Both the resulting attention distribution and the empirical \emph{importance distribution} \( I_M(D) \) are normalized. 

\subsection{Top-$k$ Ranking Alignment with Importance Scores}
\label{subsec:attention_results}

We evaluate the top-$k$ ranking consistency of attention heads by visualizing the average NDCG@10 per head across the CNN/DailyMail, SAMSum, and DECODA datasets (Figs.~\ref{fig:attention_ndcg_cnn_dailymail}--\ref{fig:attention_ndcg_decoda} in the Appendix).
To examine performance trends by depth, we also plot the average NDCG@10 across layers for each dataset (Fig.~\ref{fig:attention_ndcg_vs_layer} in the Appendix). Our analysis reveals several key observations:

\textbf{Dataset Effect.} The alignment strength varies significantly by dataset. SAMSum exhibits strong alignment, whereas CNN/DailyMail shows weak alignment. This disparity likely stems from dataset properties such as abstractiveness, document length, and summary style, which influence how well attention mechanisms encode the underlying importance ranking. The shorter, more extractive nature of SAMSum dialogues appears to align more naturally with attention patterns.

\textbf{Layer-Wise Specialization.} Performance peaks in different layers across models. While early layers show some alignment, the most proficient ranking frequently occurs in middle to late layers, indicating depth-wise functional specialization.

\subsection{MDS Visualization}
\label{subsec:attention_heads_mds_visualization}

\begin{figure}
    \centering
    \includegraphics[trim=0 0 0 3cm, clip, width=\columnwidth]{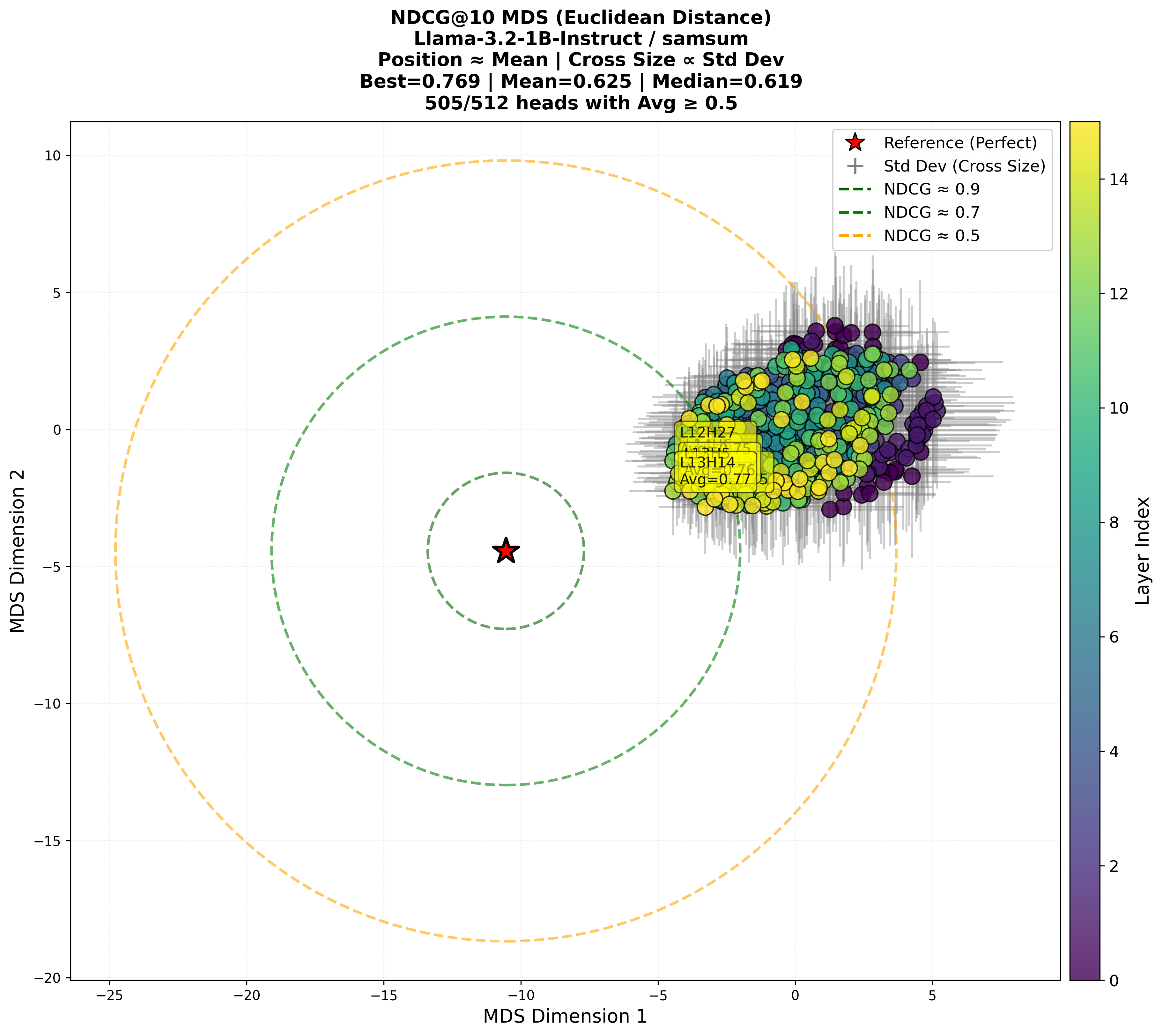}
    \caption{Multi-Dimensional Scaling (MDS) projection of attention heads for Llama-3.2-1B-Instruct on SAMSum. Points represent heads positioned by their per-sample NDCG@10 profiles, reflecting similarity in top-$k$ ranking quality with importance distribution. Point color indicates layer depth (lighter = deeper). The red star marks the ideal point of perfect ranking alignment (NDCG@10 = 1.0); dashed contours indicate similarity thresholds (e.g., $\text{NDCG@10} \geq 0.9$).}
    \label{fig:attention_mds_ndcg}
\end{figure}

We employ Multidimensional Scaling (MDS) to visualize the clustering of attention heads based on their performance profiles. Figure~\ref{fig:attention_mds_ndcg} presents this visualization for the Llama-3.2-1B-Instruct model on the SAMSum dataset, embedding each head in a two-dimensional space based on the similarity of their per-sample NDCG@10 score vectors.

The MDS projection reveals that 98.6\% of attention heads (505/512) achieve NDCG@10 $\geq$ 0.5 on SAMSum. Heads cluster progressively closer to the reference distribution (NDCG@10 = 1.0) in deeper layers, with layer 13 containing some of the highest performers (e.g., L13H14, NDCG@10 = $0.769 \pm 0.114$). The tight clustering of high-performing heads, together with small per-head standard deviations (indicated by minimal cross-markers), demonstrates that the model's top-$k$ ranking behavior is both effective and consistent across inputs.
Corresponding results using other metrics (e.g., Spearman correlation) for the best-performing head are provided in Appendix~\ref{appendix:best_heads_metrics}. 

\textbf{Overall}, these results suggest that inspection of attention distribution alone is already predictive of the distributional behavior of the models when generating summaries.

\section{Probing Experiments}
\label{sec:probing}

We train surrogate models (probes) to predict the empirical \emph{importance distribution} \( I_M(D) \) from model-internal hidden states. Each token in the input document is assigned an importance score corresponding to the average importance score of the information unit (word) to which it belongs, as defined by \( I_M(D) \).
As input to the probe, we use the hidden state vectors associated with individual tokens at a given transformer layer.  
For words that occur multiple times within a document, we aggregate their representations by averaging their hidden state vectors, yielding a single representation per word.
All probes are implemented as a one-hidden layer multi-layer perceptrons (MLPs). They were trained for 20 epochs using the Kullback–Leibler (KL) divergence loss on a 60:20:20 train/validation/test split, with early stopping (patience=3).
Additional details regarding training and dataset statistics are provided in Appendix~\ref{appendix:hs_extraction}.

\subsection{Three Probe Training Scenarios}

We define three probing scenarios to investigate which components of model $M$ encode information predictive of \emph{importance distribution} \( I_M(D) \). They are defined as follows:

\begin{enumerate}[leftmargin=*, noitemsep, topsep=3pt]
    \item \textbf{Layer-wise Probing.} A separate probe is trained on the hidden states of \emph{each individual layer} (including the embedding layer) to predict $I_i$ \emph{per token}. This isolates the predictiveness of individual hidden layers.
    \item \textbf{All-layers Probing.} A single probe is trained on features from the \emph{concatenated hidden states of a token across all layers}, predicting $I_i$ \emph{per token}. This tests if combining cross-layer information improves token-level prediction.
    \item \textbf{Article-level Probing.} A separate probe is trained \emph{per layer}, but for all tokens in parallel with a single KL loss over \emph{all tokens}. 
\end{enumerate}

\subsubsection{Probing Baselines}
\textbf{TextRank baseline.}
We compute TextRank scores as an unsupervised, content-based reference for probe evaluation.  
For each document, TextRank scores (Section~\ref{subsec:baseline_approaches}) are aligned with the empirical \emph{importance distribution} \( I_M(D) \) using the union vocabulary and normalized to sum to one.  
They are then compared against the target empirical importance distribution as a measure of the agreement between a parameter-free salience signal, providing a baseline for probe performance. 

\textbf{Randomized-weights baseline.}
To further control for false discoveries, we evaluate probing performance on models with randomized weights.  
This \textit{dead salmon} baseline tests whether predictive performance arises from pretrained representations rather than probe capacity alone on random projections of the embeddings. Following \citet{méloux2025deadsalmonsaiinterpretability}, we reinitialize all model parameters using the architecture’s native initialization scheme, extract hidden states using the same procedure as for pretrained models, and train probes to predict \( I_M(D) \).

Performance is reported on the test set using Spearman’s rank correlation and NDCG@10. Further results for all model--dataset pairs are reported in Appendix Tables~\ref{tab:textrank_spearman_baselines} and~\ref{tab:textrank_ndcg_baselines}.

\subsection{Experimental Results for Article-level Probing}
\label{sec:probing_experimental_results}

This section presents the results for \textbf{Scenario 3: Article-level Probing}, where the probing loss is aggregated across all tokens in a document. This evaluates the probe's ability to reconstruct the relative importance distribution for an entire document context.
Results for \textbf{Scenario 1 (Layer-wise Probing)} and \textbf{Scenario 2 (All-layers Probing)} are provided in Appendix~\ref{appendix:layer_wise_probing} and Appendix~\ref{appendix:all_layers_probing}, respectively.

\begin{figure*}[!htb]
\centering
  \includegraphics[trim=0 0 0 0cm, clip, width=.9\textwidth]{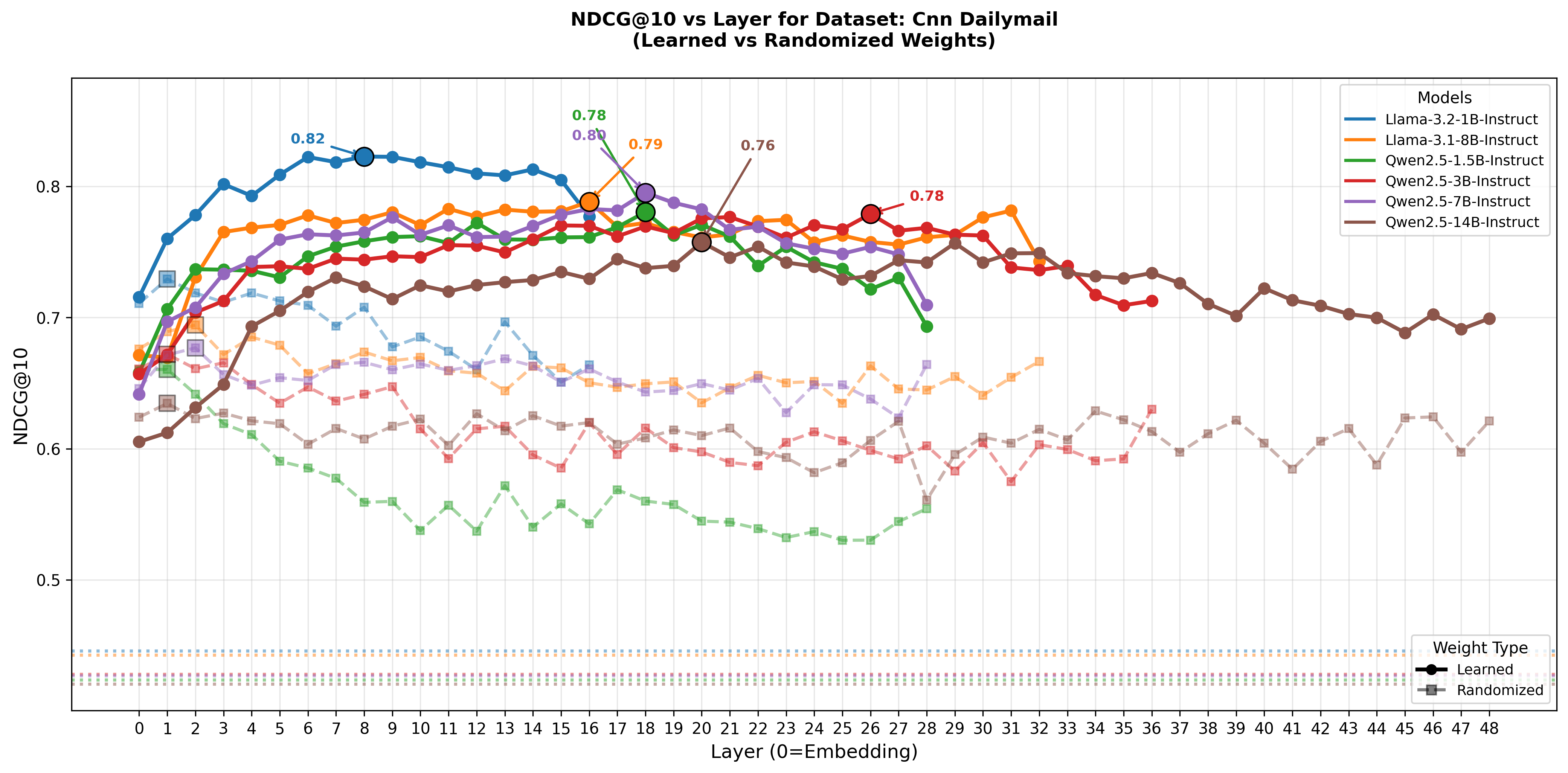}
  \includegraphics[trim=0 0 0 0cm, clip, width=.9\textwidth]{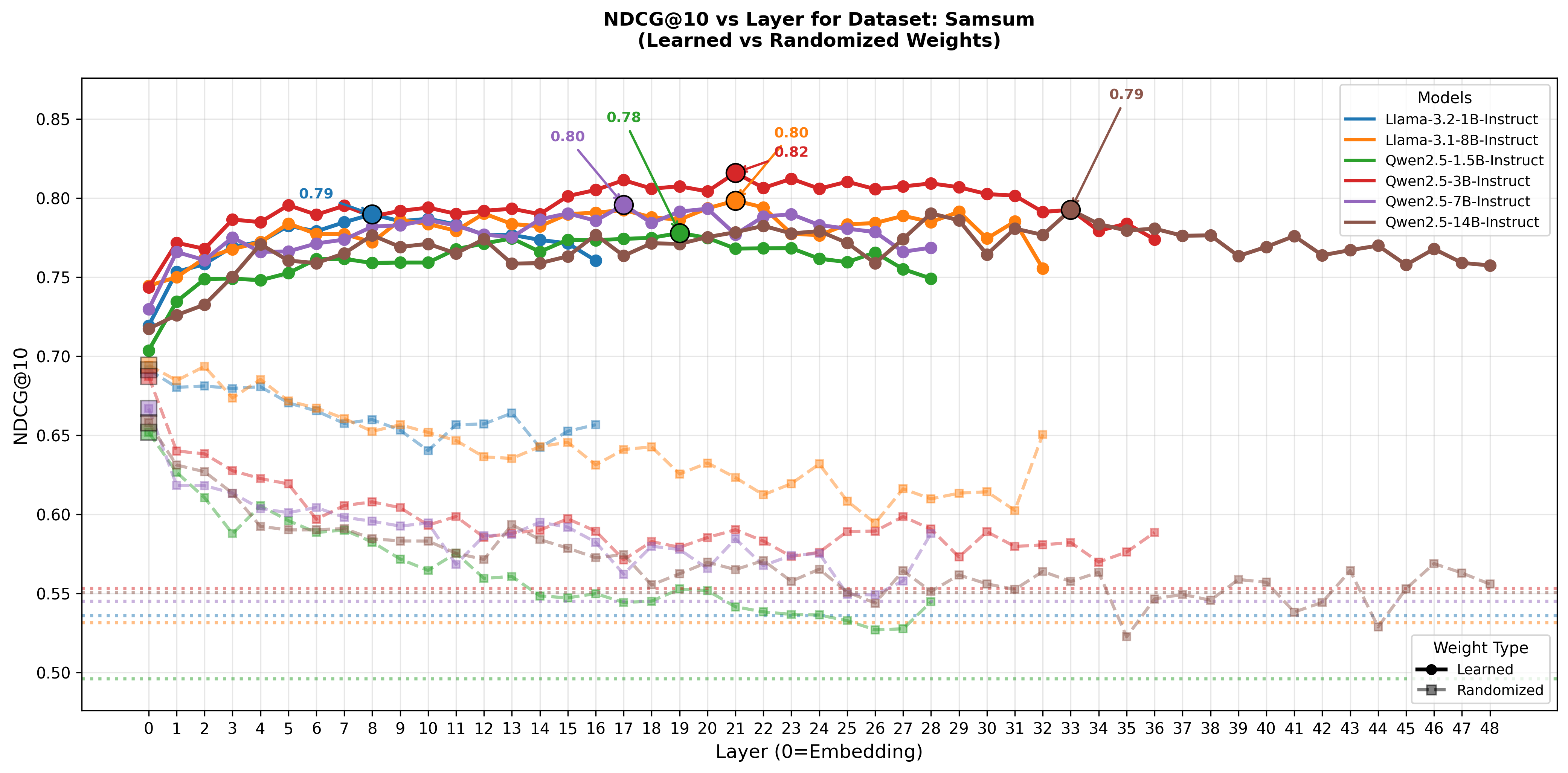}
    \caption{Article-level probing NDCG@10 across layers for CNN/DailyMail (top) and SAMSum (bottom). Round dots show learned model performance; square dots show the \textit{Randomized Weights Baseline}. The best-performing layers are annotated. Horizontal dashed lines show the TextRank baseline for each model.}
    \label{fig:S3_article_level_ndcg}
\end{figure*}

\begin{figure*}[!htb]
\centering
  \includegraphics[width=.9\textwidth]{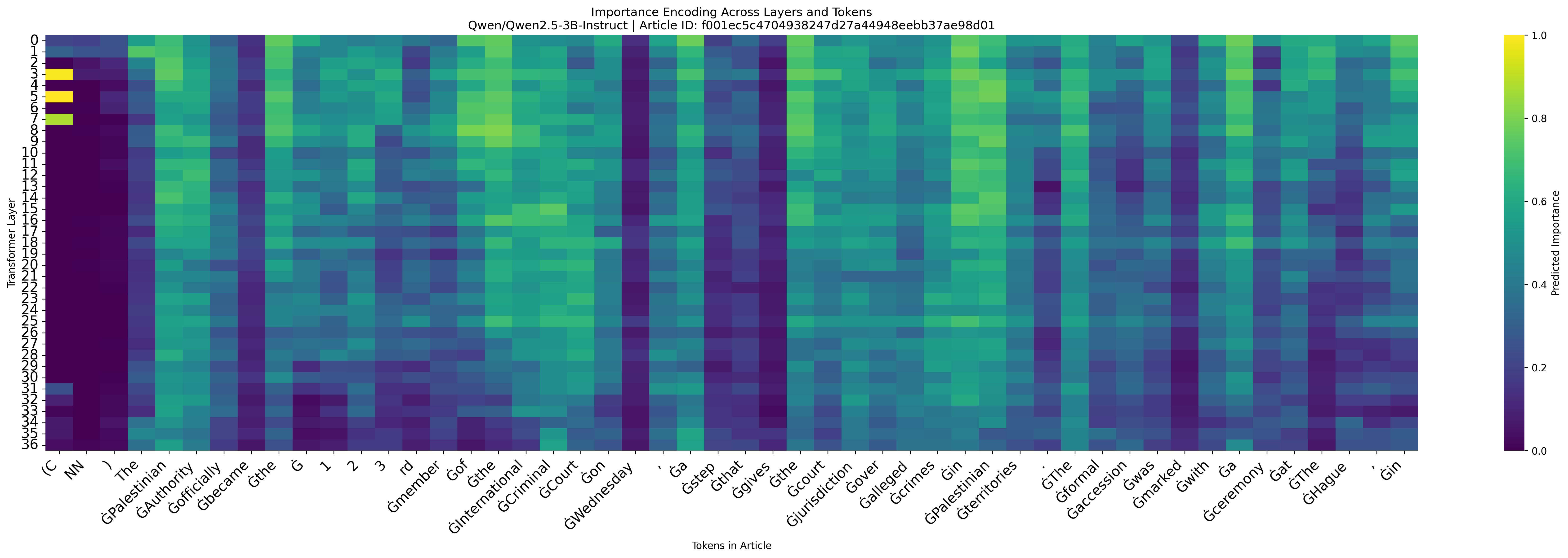}
    \includegraphics[width=.9\textwidth]{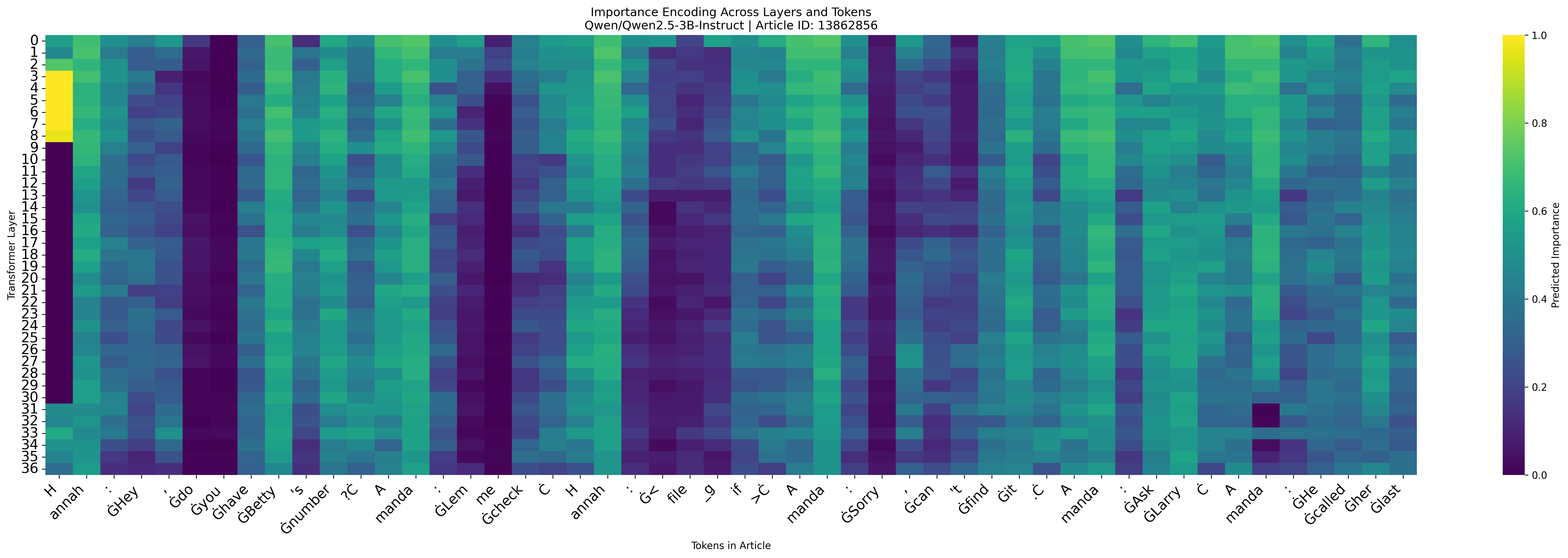}
  \caption{Heatmap visualization of layer-wise importance predictions for Qwen2.5-3B-Instruct, showing probe outputs across all layers and the first 50 tokens for representative samples (top: CNN/DailyMail, bottom: SAMSum).}
  \label{fig:sample_analysis}
\end{figure*}

\subsubsection{Quantitative Results}
\label{subsec:probe_quantitative_results}
This section reports NDCG@10 results on CNN/DailyMail and SAMSum datasets (Figure~\ref{fig:S3_article_level_ndcg}); complementary Spearman results are provided in Appendix~\ref{appendix:article_level_probing}. 
NDCG@10 values are generally high, indicating that probes capture locally consistent word-importance rankings. Performance rises in early layers and peaks in middle-to-late layers across all datasets, suggesting that the most discriminative information for top-$k$ ranking is encoded at intermediate depths. Unlike Spearman correlation, which remains stable on SAMSum and DECODA, NDCG@10 exhibits pronounced layer-wise variation across all three datasets.

Probes trained on learned model weights substantially outperform those trained on randomized-weight models, achieving improvements of 0.1–0.2 NDCG@10 points, and consistently exceed the TextRank baseline. In contrast, randomized-weight probes show degraded performance and high inter-layer variance. This demonstrates that the LLMs do indeed construct representations that are predictive of the importance distributions.

\subsubsection{Example Analysis}

We select Qwen2.5-3B-Instruct for a case study because of its balanced performance: it achieves the highest NDCG@10 score on SAMSum while ranking in the middle on CNN/DailyMail (Section~\ref{subsec:probe_quantitative_results}). Figure~\ref{fig:sample_analysis} visualizes article-level probe predictions via heatmaps, illustrating the layer-wise encoding of importance for representative samples from both datasets.
The heatmaps (Fig.~\ref{fig:sample_analysis}) plot predicted importance across transformer layers (vertical axis) versus the first 50 input tokens (horizontal axis), with intensity indicating encoding strength.
Token-level analysis reveals consistent high importance across layers for: (1) CNN/DailyMail: named entities (e.g., ``Palestinian'', ``Authority'', ``the international criminal court''); (2) SAMSum: person names (e.g., ``Hannah'', ``Betty'', ``Amanda'', ``Larry'') and core dialogue tokens (``number'', ``it'', ``her'').
These patterns reflect dataset characteristics: \emph{importance} is broadly distributed across content in news articles, but focuses more heavily on participant references in dialogues.

\section{Discussion and Conclusion}

This study provides a computational and behavioral investigation into the latent \textit{information importance} that guides content selection in LLMs. We derived an empirical importance distribution \( I_M(D) \) by analyzing information persistence across length-controlled summaries. Our behavioral analysis shows that LLMs display consistent importance patterns distinct from pre-LLM baselines, with models clustering more strongly by architecture than by size.

To predict \( I_M(D) \), we employed attention analysis and hidden-state probing. Both approaches identify the middle-to-late layers as critical for encoding importance, with specific attention heads serving as effective predictors. These results show that inspection of the internal computation of the models while processing a document can be highly predictive of the distributional importance of information units. However, the strength of the predictions is dataset-dependent (strongest on SAMSum, weakest on CNN/DailyMail) and model-dependent (with Llama models outperforming Qwen).
Our central findings are twofold. First, the encoding of importance is \textbf{insensitive to model scale}, showing no consistent improvement with increased parameters. Second, it is \textbf{highly task-specific}, being far more predictable in conversational dialogues (SAMSum) than in long-form news (CNN/DailyMail). This suggests that a model's capacity for importance ranking is less a function of its size and more a product of its architectural design, training data and the inherent nature of the source text.

Our work provides initial insights into LLMs' summarization priorities and their internal representations, paving the way toward interpreting and controlling information selection. A key gap remains: while models maintain consistent internal hierarchies, we lack direct methods to access or control them. Future work should bridge this gap through causal manipulation experiments for output control. 

\section*{Limitations}

This study has several limitations. First, our data processing pipeline involves numerous choices (e.g., handling of function words) that could be refined to ensure high importance scores correspond more directly to substantive content.
Second, despite using length-constraint prompts, not all generated summaries strictly adhered to the specified token counts, introducing variance.
A third limitation stems from the data characteristic detailed in Appendix~\ref{appendix:hs_extraction}: the imperfect alignment between annotated summary words and source tokens results in a limited set of word--hidden-state pairs for probe training. Future work could improve this mapping by incorporating semantic similarity, moving beyond exact lexical matching. In general, other information units more semantically meaningful can be inspected in the future. 

\section*{Acknowledgments}
This work was partially supported by the ``Intelligent Systems for Data, Knowledge, and Humans'' axis of the Grenoble Computer Science Laboratory (LIG).
This work was conducted within the French research unit UMR 5217 and was supported by CNRS (grant ANR-22-CPJ2-0036-01 and ANR-25-CE23-2059-01) and by MIAI@Grenoble-Alpes (grant ANR-19-P3IA-0003 and ANR-23-IACL-0006).

\bibliography{custom,anthology}

\appendix

\section{Multilingual Study}
\label{appendix:french_decoda}

To assess the cross-lingual generalizability of our findings, we extend our evaluation to a French dialogue summarization dataset, complementing the English datasets described in Section~\ref{subsubsec:datasets}.

We employ the \textbf{DECODA} corpus \citep{favre-etal-2015-call}, a French call center dialogue collection created for the Multiling 2015 CCCS abstractive summarization task. The corpus contains 1,000 unannotated dialogues and a test set of 100 dialogues with human-written synopses. We use the preprocessed samples from \citet{zhou-etal-2022-effectiveness}, which retain speaker identifiers while removing extraneous labels.

All reported results, including behavioral analysis, attention alignment, and article-level probing (Scenario~3), are based on the 100-dialogue test set. These results are presented in the following sections alongside supplementary results for the CNN/DailyMail and SAMSum datasets.

\section{Metrics for Word Importance Evaluation}
\label{appendix:metrics}

Evaluating word importance attribution is non-trivial due to the lack of a standard metric for comparing \emph{importance distribution} \( I_M(D) \), unlike for generated text (e.g., ROUGE, BERTScore). A core challenge is the absence of a clear theoretical baseline, which complicates both metric selection and the interpretation of results.

\subsection{Candidate Metrics}
To rigorously assess the alignment between model-generated \emph{importance distribution} \( I_M(D) \) and a human gold standard (derived from reference summaries), we evaluate a comprehensive suite of metrics spanning four categories:
\begin{itemize}
    \item \textbf{Correlation}: Spearman's $\rho$ and Kendall's $\tau$, measuring global rank alignment.
    \item \textbf{Ranking}: Normalized Discounted Cumulative Gain (NDCG@$k$), Precision@$k$, and Recall@$k$, focusing on the retrieval of top-important words.
    \item \textbf{Distributional Divergence}: Kullback-Leibler (KL) Divergence, Jensen-Shannon Divergence (JSD), and Rényi Divergence, measuring the distance between probability distributions.
    \item \textbf{Set Overlap}: Jaccard Similarity, measuring the intersection of selected important word sets.
\end{itemize}

\subsection{Meta-Evaluation Methodology}
Given the diversity of available metrics, we perform a meta-evaluation to identify the most robust indicators of quality. We evaluate each metric based on two key criteria:
\begin{enumerate}
    \item \textbf{Discrimination Power ($D$)}: The standard deviation of the metric scores across different models. A higher $D$ indicates the metric effectively distinguishes between models of varying quality.
    \item \textbf{Sensitivity ($S$)}: The ratio of the observed range of scores to the theoretical range of the metric. An ideal sensitivity (approaching 1.0) implies the metric utilizes its full scale and is not saturated.
\end{enumerate}
We compute a \textbf{Composite Score} for each metric, defined as the average of its normalized \textit{Discrimination} and \textit{Sensitivity} scores, to rank their overall empirical utility.

\subsection{Empirical Results}
Our meta-evaluation on the CNN/DailyMail (Long-form news, 3,000 samples) and SAMSum (Short-form dialogue, 819 samples) datasets yielded the following insights:

\paragraph{CNN/DailyMail Results}
The divergence-based metrics demonstrated high discrimination power. \textbf{Rényi Divergence ($\alpha=2.0$)} achieved the highest composite score (0.864), followed by \textbf{KL Divergence} (0.748). Among bounded metrics, \textbf{Spearman's $\rho$} ranked third (0.746) with high sensitivity (0.86), and \textbf{NDCG@10} ranked fifth (0.704) with perfect sensitivity (1.00).

\paragraph{SAMSum Results}
For the dialogue dataset, overlap-based metrics performed exceptionally well. \textbf{Jaccard@15} achieved the highest composite score (0.965), likely due to the shorter, more keyword-centric nature of the dialogues. However, \textbf{Rényi Divergence ($\alpha=2.0$)} remained robust (3rd, 0.816), and \textbf{NDCG@10} (4th, 0.754) and \textbf{Spearman's $\rho$} (8th, 0.706) continued to show strong performance with perfect sensitivity (1.00).

\subsection{Choice of Metrics}
Based on the empirical results and theoretical properties, we select \textbf{Spearman's Rank Correlation ($\rho$)} and \textbf{NDCG@10} as our primary universal metrics for the following reasons:

\begin{enumerate}
    \item \textbf{Complementarity}: Spearman captures the \textit{global} alignment of the entire importance distribution, while NDCG@10 focuses specifically on the \textit{local} quality of the most important words (top-10), which are most critical for summarization.
    \item \textbf{Robustness}: Both metrics demonstrated high sensitivity ($>0.85$) across both datasets, ensuring they remain meaningful regardless of the data domain (news vs. dialogue).
\end{enumerate}

Although Rényi Divergence ($\alpha=2.0$) demonstrated high discriminative power, we reserve it as a secondary diagnostic tool due to its unbounded range, prioritizing the bounded and more interpretable Spearman $\rho$ and NDCG@10 for our primary analysis.

\section{Details on the Experimental Setup}

\subsection{Prompts for Data Generation}
The prompts used for generating length-variant summaries across datasets are presented in Table \ref{tab:prompt_data_generation}, where $N \in \{10, 20, 30, \dots, 100\}$.

\begin{table}[htbp]
\centering
\begin{tcolorbox}[
  colback=myBlue!5,
  colframe=myBlue!75,
  title=Prompt for Data Generation,
  fonttitle=\bfseries,
  ]
  
\small
\textbf{CNN/DailyMail}: \\
``Summarize the following text in exactly \{N\} words: \{text\}''

\textbf{SAMSum}: \\
``Summarize the following dialogue in exactly \{N\} words: \{dialogue\}''

\textbf{DECODA}: \\
``Résumez le dialogue suivant en exactement \{N\} mots: \{dialogue\}''

\end{tcolorbox}
\caption{Prompts used for generating length-variant summaries across datasets.}
\label{tab:prompt_data_generation}
\end{table}

\subsection{Model Specification}
\label{appendix:model_specification}

In Table \ref{tab:model_specification}, we list the models and their corresponding links.

\begin{table*}[!htp]
  \centering
  \small
  \begin{tabular}{ll}
    \hline
    \textbf{Models}           & \textbf{Links}  \\
    \hline
    Llama-3.2-1B-Instruct \citep{llamamodelscard}      & \url{https://huggingface.co/meta-llama/Llama-3.2-1B-Instruct}            \\
    Llama-3.1-8B-Instruct   \citep{llamamodelscard}  & \url{https://huggingface.co/meta-llama/Llama-3.1-8B-Instruct}           \\
    Qwen2.5-1.5B-Instruct \citep{qwen2.5}      & \url{https://huggingface.co/Qwen/Qwen2.5-1.5B-Instruct}              \\
    Qwen2.5-3B-Instruct \citep{qwen2.5} & \url{https://huggingface.co/Qwen/Qwen2.5-3B-Instruct} \\
    Qwen2.5-7B-Instruct \citep{qwen2.5} & \url{https://huggingface.co/Qwen/Qwen2.5-7B-Instruct}      \\
    Qwen2.5-14B-Instruct \citep{qwen2.5} & \url{https://huggingface.co/Qwen/Qwen2.5-14B-Instruct} \\
    deepseek-chat$^\dag$  \citep{deepseekai2025deepseekv3technicalreport}  &        \url{https://api-docs.deepseek.com/}                    \\
    \hline
  \end{tabular}
  \footnotesize{
  $^\dag$Experiments done in September 2025, the model points to non-thinking mode of DeepSeek-V3.1-Terminus.}
  \caption{\label{tab:model_specification}
    Models used in the experiments and their corresponding links.
  }
\end{table*}

\subsection{Analysis of Importance Score Distributions}
\label{appendix:importance_score_analysis}

\begin{figure*}[!htb]
\centering
    \includegraphics[width=\textwidth]{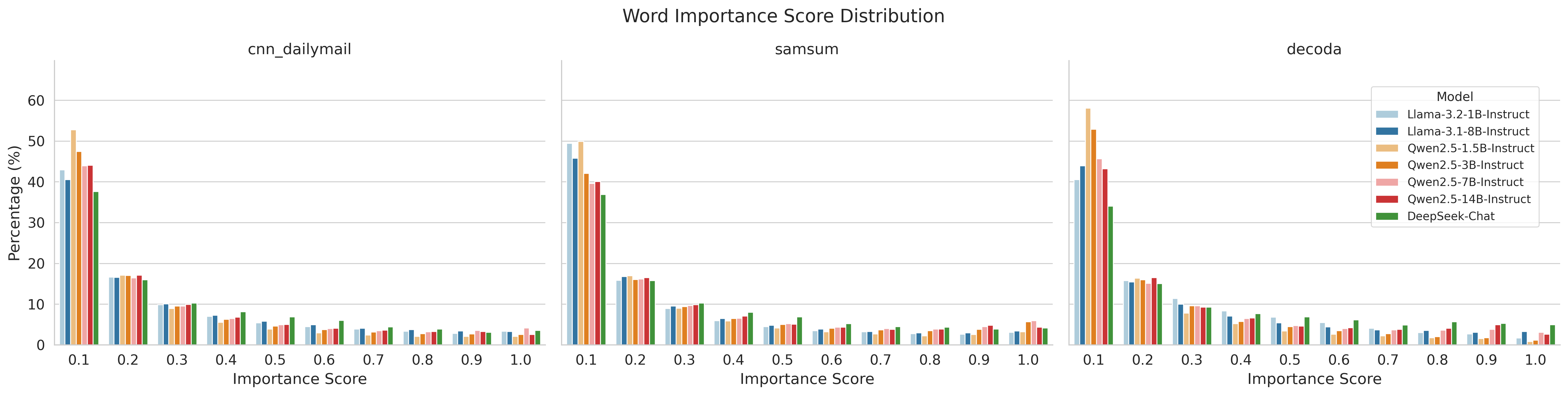}
    \caption{Distribution of word importance scores across models and datasets.}
  \label{fig:importance_score_grouped_bar}
\end{figure*}

Figure \ref{fig:importance_score_grouped_bar} presents the distributions of importance scores across all models for the three datasets. The distributions reveal that the majority of words receive low importance scores, with approximately 50\% of all word-importance pairs assigned a score of 0.1 across all three datasets. Only a small fraction ($\sim$6-8\%) of words receive scores $\geq$0.8.

\paragraph{CNN/DailyMail}

For the CNN/DailyMail dataset, the derived importance scores show a hierarchy: the highest scores ($\geq 0.9$) are assigned to \textbf{central content words}, including named entities, main events. Medium scores correspond to \textbf{supporting information}, while the lowest scores ($\leq 0.2$) are given to function words, stop words, and peripheral background details.

All evaluated models demonstrate strong agreement on the most important words, consistently identifying main subjects, verbs, and unique entities. Substantial overlap occurs in the top 10–20\% of important words across models, with greater variation appearing in the middle and low importance ranges. For example:

\begin{itemize}[noitemsep,topsep=0pt,parsep=0pt,partopsep=0pt]
    \item High Importance ($\geq$0.9): ``international'', ``authority'', ``court'', ``criminal'', ``palestinian'', ``state'', ``stray'', ``washington'', ``dog'', ``in'' (central entities and concepts).
    \item Medium Importance (0.4–0.6): ``peace'', ``marking'', ``into'', ``investigate'', ``over'', ``jurisdiction'', ``on'', ``as'', ``field'' (supporting details and context).
    \item Low Importance ($\leq$0.2): ``joins'', ``formal'', ``with'', ``giving'', ``face'', ``but'', ``if'', ``be'', ``its'', ``whether'' (function words, connectors, and background terms).
\end{itemize}

\paragraph{SAMSum}
The SAMSum dataset shows similar patterns, with \textbf{named entities and main actors} receiving the highest importance scores, while function words and background terms are consistently assigned low importance. For example:

\begin{itemize}[noitemsep,topsep=0pt,parsep=0pt,partopsep=0pt]
    \item High Importance ($\geq$0.9): ``amanda'', ``find'', ``eric'', ``rob'', ``lenny'', ``bob'' (named entities and key actors).
    \item Medium Importance (0.4–0.6): ``for'', ``if'', ``ask'', ``phone'', ``routine'' (supporting details and context).
    \item Low Importance ($\leq$0.2): ``tries'', ``texting'', ``exchange'', ``so'', ``suggesting'', ``do'', ``of'', ``goodbye'' (function words, connectors and supplementary verbs).
\end{itemize}

\paragraph{DECODA}
The DECODA dataset (French customer service dialogues) exhibits a distinct pattern where the model prioritizes both \textbf{domain-specific content} and essential \textbf{grammatical markers}. Unlike the English datasets, determiners and prepositions frequently receive high importance scores, likely due to their critical role in French syntax and coreference resolution.

\begin{itemize}[noitemsep,topsep=0pt,parsep=0pt,partopsep=0pt]
    \item High Importance ($\geq$0.9): ``numéro'', ``métro'', ``bus'', ``client'', ``remboursement'' (domain entities), along with ``le'', ``un'', ``pour'', ``de'' (determiners/prepositions).     
    \item Medium Importance (0.4–0.6): ``l'agent'', ``rappeler'', ``après'', ``donc'', ``faire'' (procedural verbs and connectors).
    \item Low Importance ($\leq$0.2): ``conversation'', ``échange'', ``précise'', ``bonjour'', ``est'', ``on'', ``il'' (meta-dialogue descriptors, greetings, and generic auxiliary verbs).
\end{itemize}

These patterns demonstrate that while models consistently prioritize named entities and main actions, the treatment of function words varies by language, with French models retaining more grammatical structure in the high-importance tier.

\subsection{Hidden States Extraction}
\label{appendix:hs_extraction}

\begin{table}[!htb]
\centering
\small
\include{tables/hs_statistics}
\caption{Statistics on hidden states extraction in different models on both datasets.}
\label{tab:hs_statistics}
\end{table}

Table \ref{tab:hs_statistics} presents detailed statistics on the ratio of zero-score words to annotated words. The annotated words are identified based on their frequency in the generated summaries and their presence in the corresponding article or dialogue.

The analysis encompasses 300 articles from the CNN/DailyMail dataset, with approximately 88K-95K total words per model, as well as 819 dialogues from the SAMSum dataset, containing approximately 60K-67K total words per model.

The statistics reveal a distinct pattern in hidden-state extraction between datasets. For the conversational SAMSum dataset, models identified a majority of source words as annotated (47.8\%–77.7\%), with a minority as zero-score words (22.3\%–52.2\%). The opposite trend holds for the news-based CNN/DailyMail dataset, where zero-score words constitute the majority (53.5\%–65.3\%) and annotated words are the minority (34.7\%–46.5\%).

This suggests that summarization operates differently by genre. In long-form news, the task is inherently more abstractive, requiring the distillation of many source words into a concise summary, which results in a high proportion of zero-score words. In shorter dialogues, summaries are more extractive, preserving a higher density of the original words. Consequently, the informative features captured by hidden states are distributed differently, with conversational data exhibiting a higher concentration of annotated, salient tokens.

\section{Behavioral Analysis}
\label{appendix:behavioral_analysis}

\subsection{Model Similarity}

\begin{figure}[!htb]
  \centering
    \includegraphics[width=\linewidth]{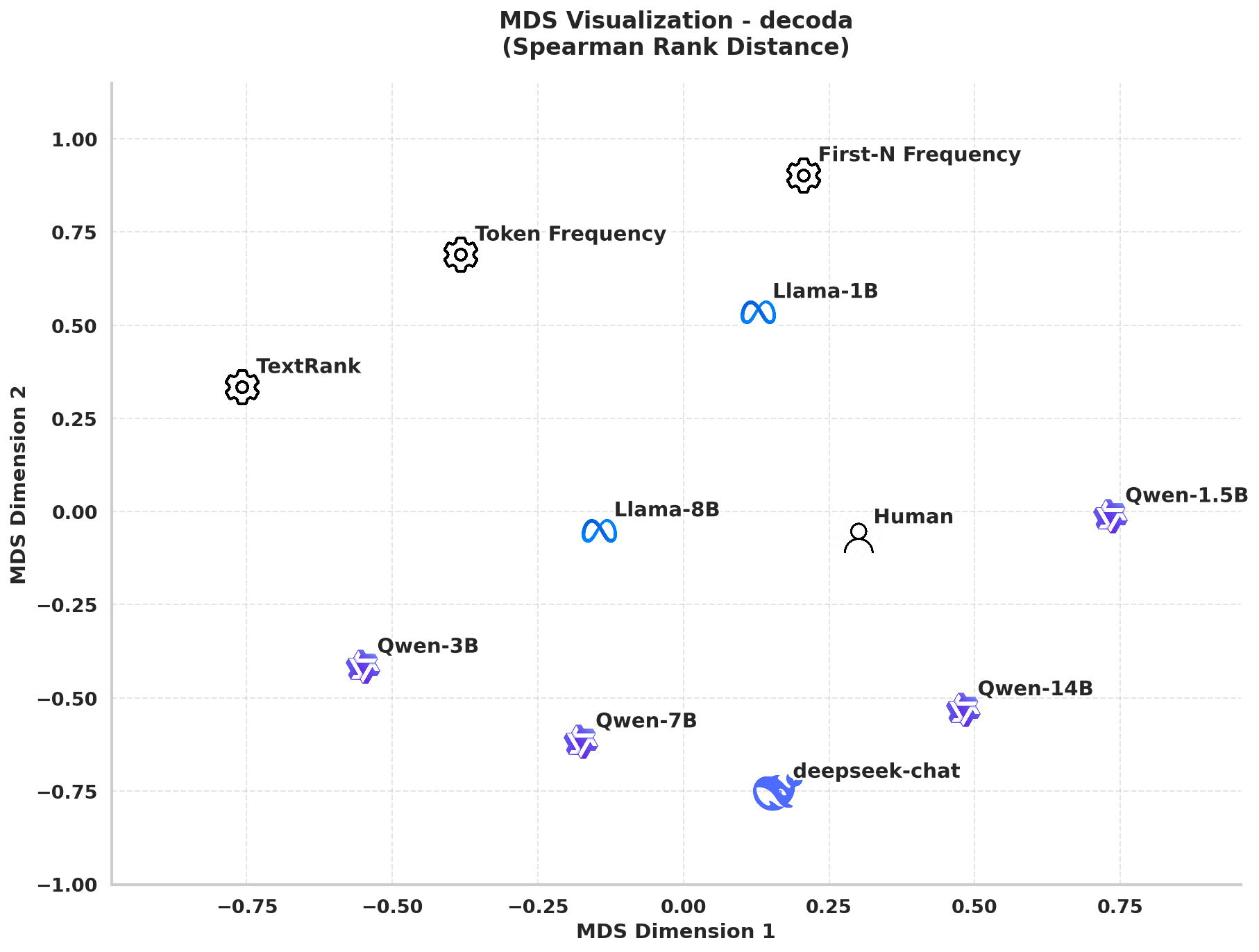}
  \caption{Pairwise model similarity based on Spearman rank correlation distance for importance distributions, visualized via two-dimensional Multidimensional Scaling (MDS). Results are shown for the DECODA dataset.} 
  \label{fig:model_similarity_spearman_decoda}
\end{figure}

\begin{figure}[!htb]
  \centering
  
  \begin{subfigure}{\columnwidth}
    \centering
    \includegraphics[width=\linewidth]{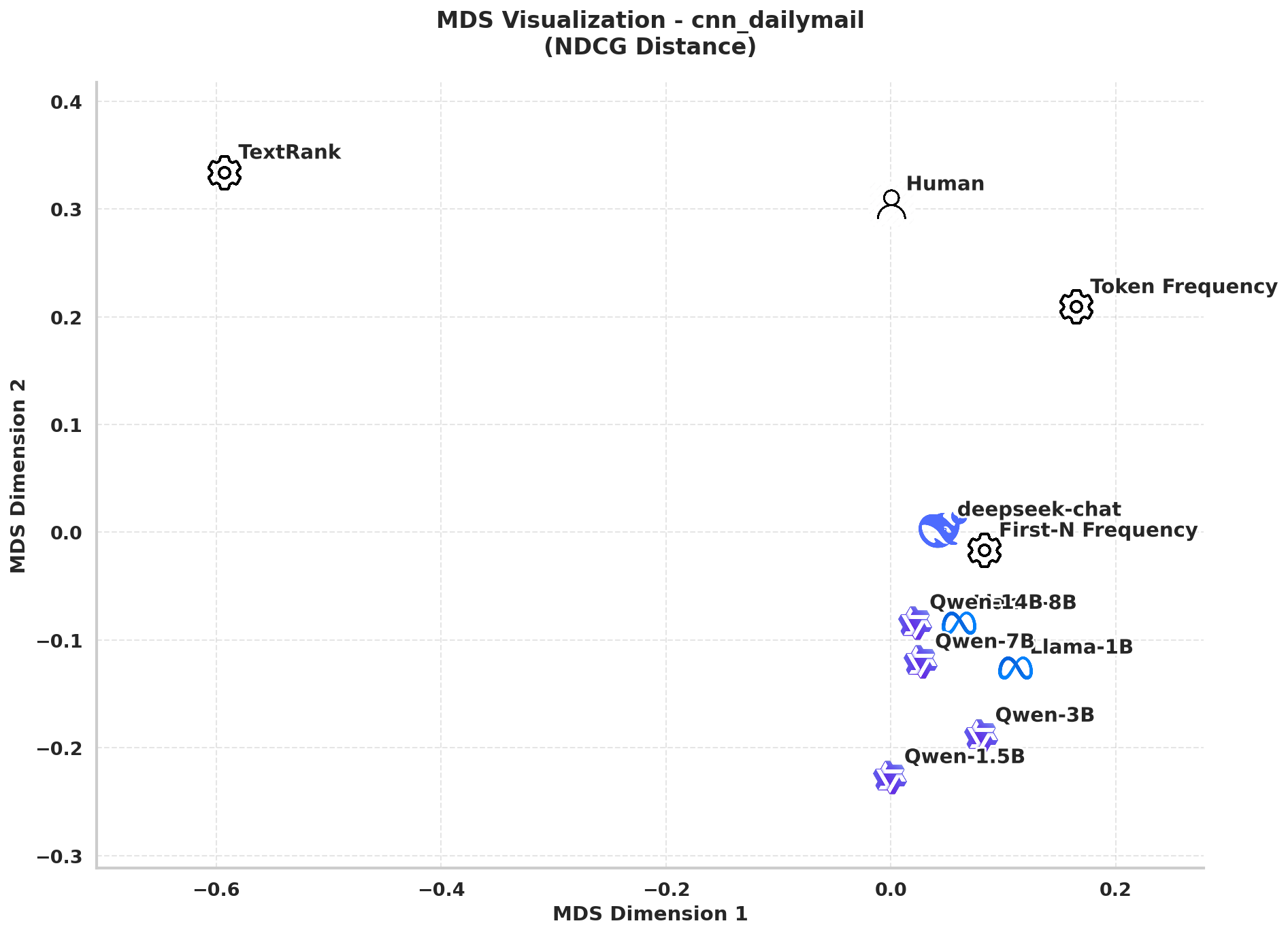}
  \end{subfigure}
  
  \vspace{0.5cm}
  
  \begin{subfigure}{\columnwidth}
    \centering
    \includegraphics[width=\linewidth]{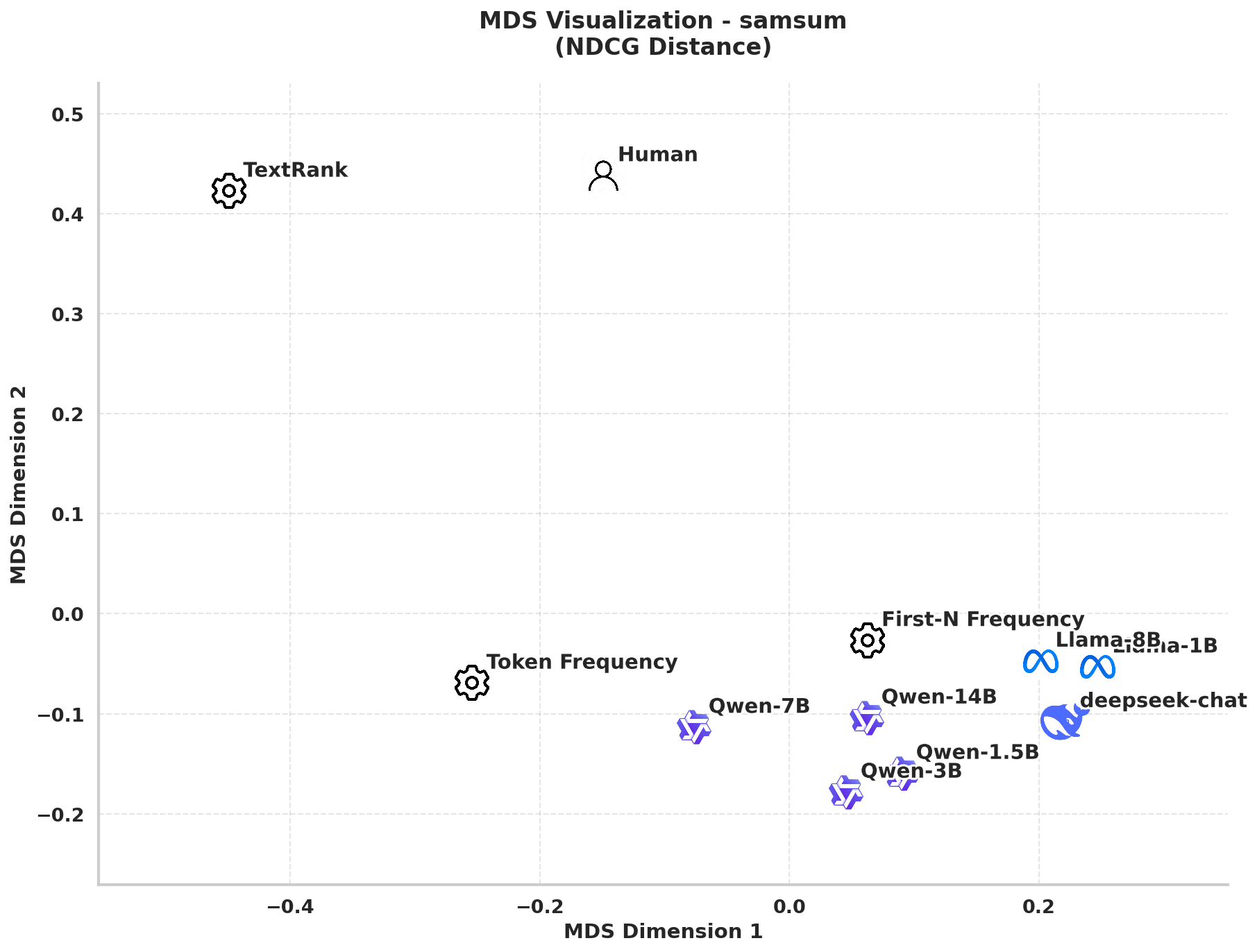}
  \end{subfigure}
  
  \vspace{0.5cm}
  
  \begin{subfigure}{\columnwidth}
    \centering
    \includegraphics[width=\linewidth]{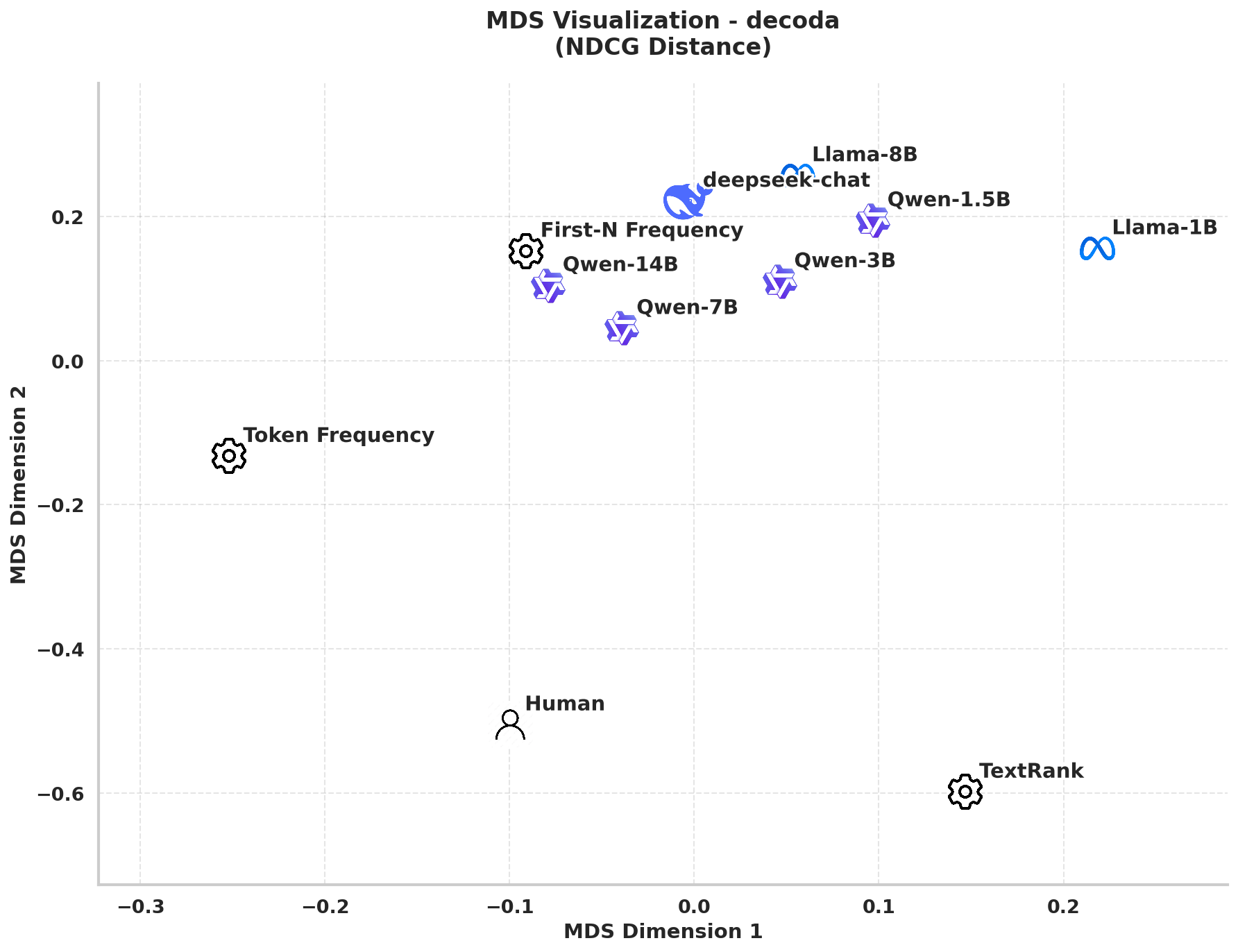}
  \end{subfigure}
  \caption{Model similarity in importance distribution rankings across the CNN/DailyMail, SAMSum, and DECODA datasets, visualized using pairwise NDCG@10 distances.}
  \label{fig:model_similarity_ndcg}
\end{figure}

We extend the model similarity analysis from Section~\ref{subsec:model_similarity} by including results from the French DECODA dataset (Figure~\ref{fig:model_similarity_spearman_decoda}) and by evaluating similarity using NDCG@10 (Figure~\ref{fig:model_similarity_ndcg}).
The latter figure visualizes pairwise model distances based on NDCG@10 across all three datasets.

The key observations from Section~\ref{subsec:model_similarity} are reinforced and sharpened using NDCG@10: the behavioral distinction between LLMs and pre-LLM baselines (excluding \textit{First-N-Frequency}) is more pronounced; clustering by model family becomes more visually distinct; and the \textit{Human (Frequency)} consistently occupies an intermediate position between the pre-LLM baselines and the LLM cluster.
These consistent patterns across two different metrics and three datasets strengthen the conclusion that LLMs share a common, family-influenced approach to attributing \emph{importance distribution} \( I_M(D) \) that differs from classical methods.

\subsection{Quantifying Positional Bias}
\label{appendix:positional_bias}

\begin{figure}[!htb]
  \includegraphics[trim=0 0 0 16.5cm, clip, width=\columnwidth]{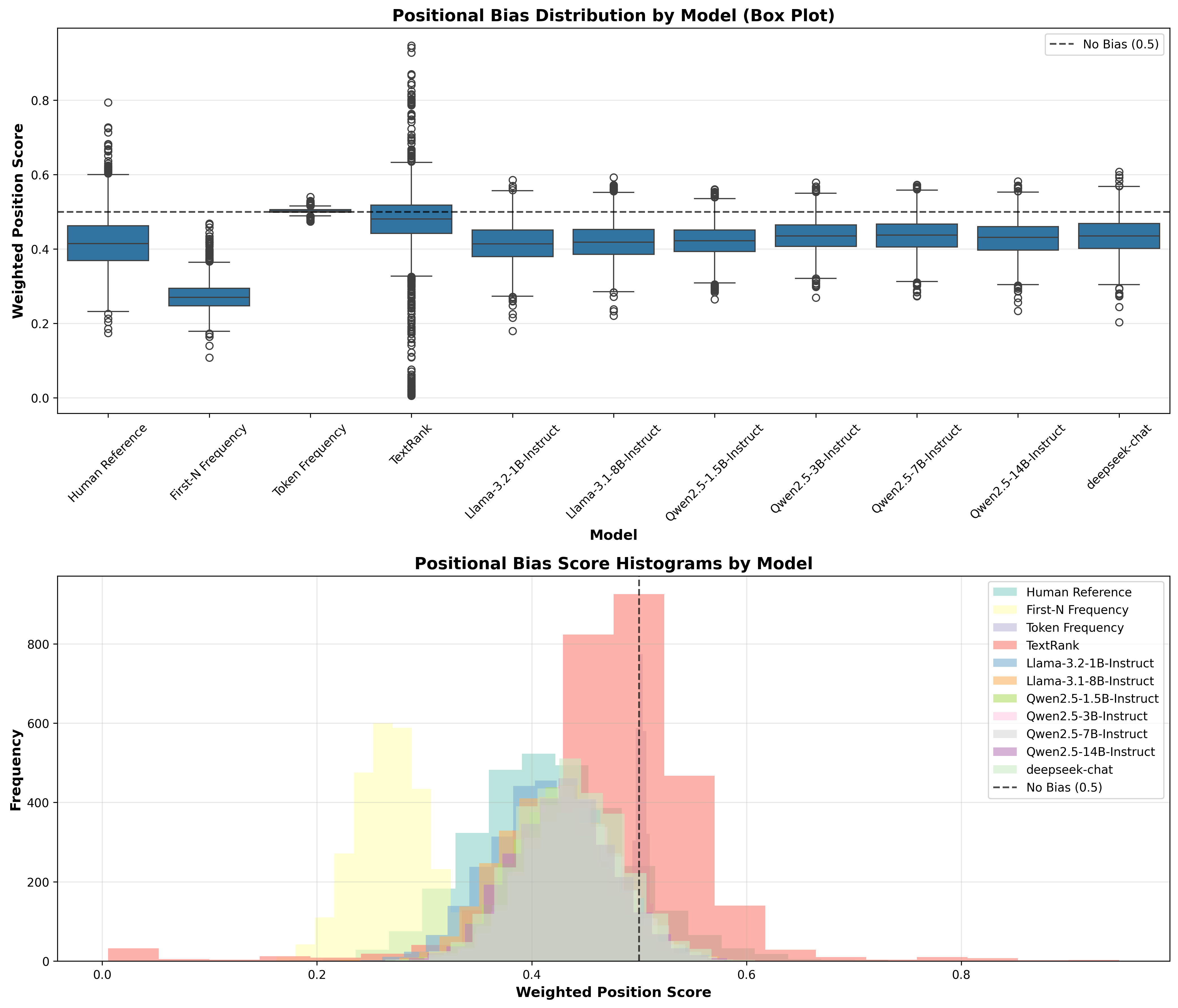} \\
 \includegraphics[trim=0 0 0 16.5cm, clip, width=\columnwidth]{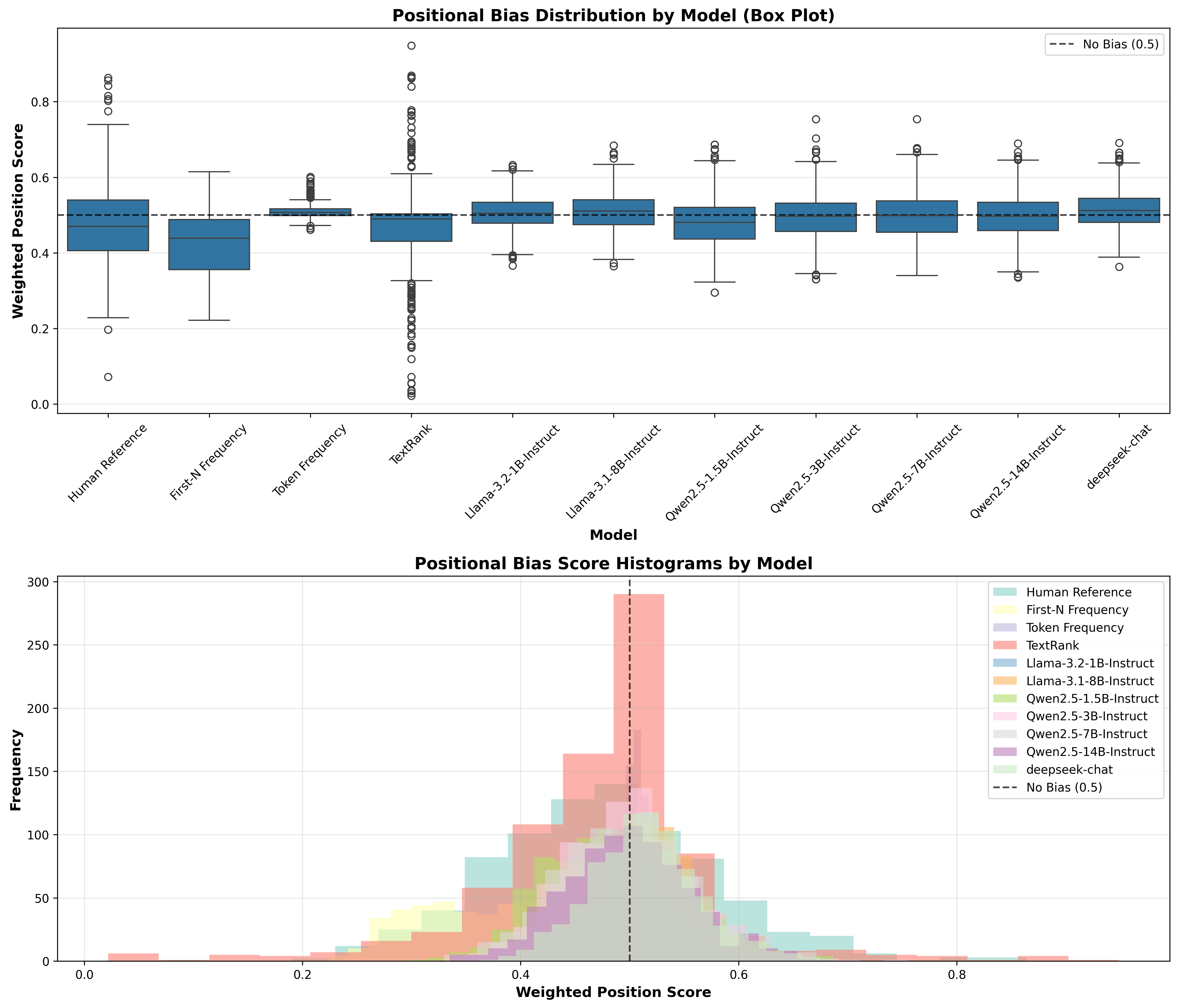}
   \includegraphics[trim=0 0 0 16.5cm, clip, width=\columnwidth]{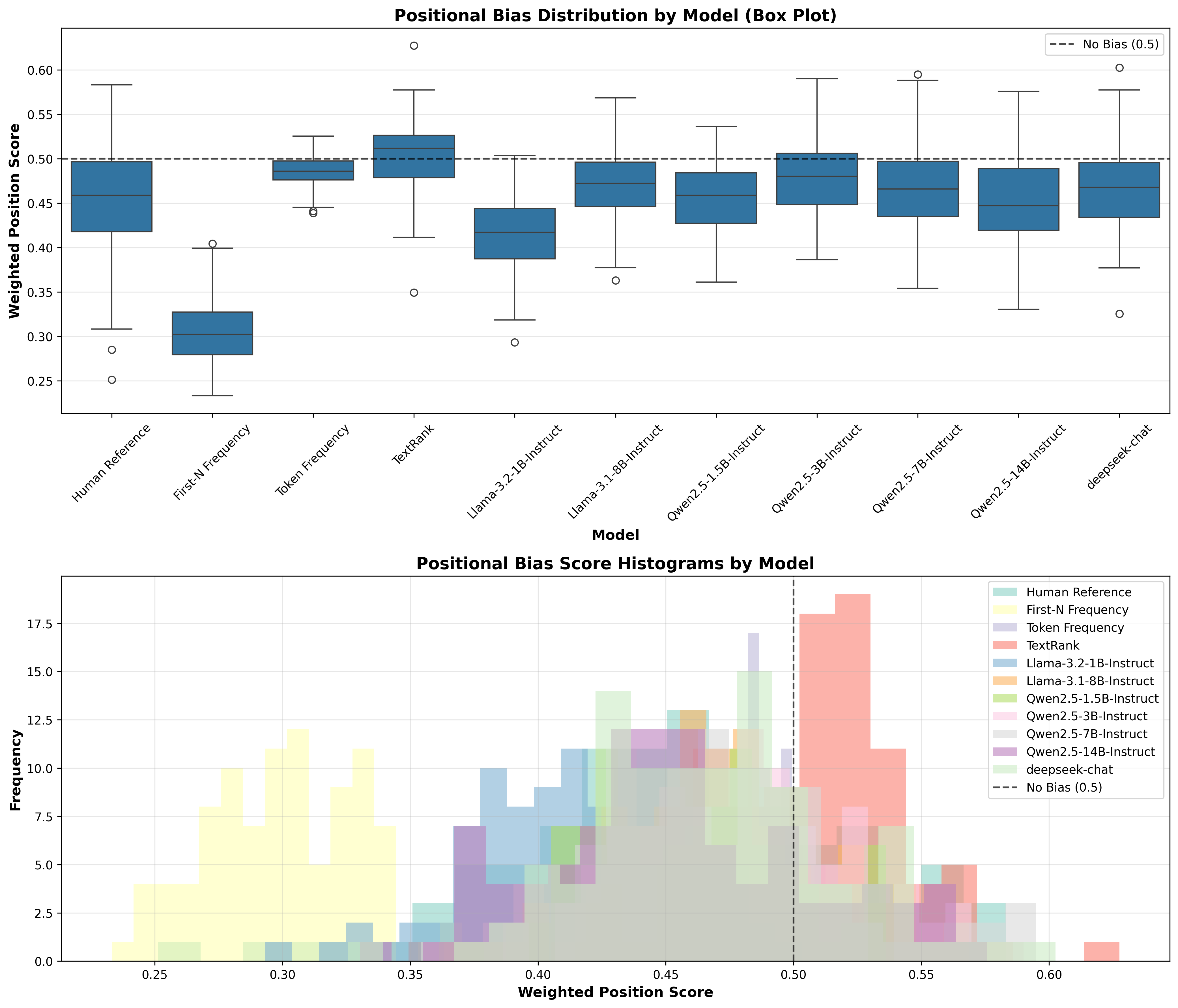} \\
  \caption{Distribution of positional bias scores for all models across the CNN/DailyMail (top), SAMSum (middle), and DECODA (bottom) datasets. Lower scores indicate a stronger bias towards earlier document positions.}
  \label{fig:positional_bias_distributions}
\end{figure}

Since human tend to favor early information (e.g., historically strong baselines used the first sentences), we quantify this positional bias using a weighted positional importance score. For each model, the bias for a document $D$ is defined as:
\[
\text{Bias}(D) = \frac{1}{S_I} \sum_{t=1}^{L} I(t) \cdot p(t), \quad \text{where } p(t) = \frac{t}{L}.
\]
Here, $I(t)$ is the importance score of token $t$, $L$ is the document length in tokens, $S_I = \sum_{t} I(t)$ is the total importance, and $p(t)$ is the normalized position (ranging from 0 at the start to 1 at the end). For words with multiple occurrences, we use the average positional index. A \emph{lower} $\text{Bias}(D)$ value indicates a stronger preference for earlier tokens.

Figure~\ref{fig:positional_bias_distributions} shows the bias distributions for each model on the three datasets. Consistent patterns emerged: the baseline \textit{First-N Frequency} exhibited the strongest early bias (lowest scores: 0.228 on CNN/DailyMail, 0.079 on SAMSum, 0.196 on DECODA), while other baselines (\textit{TextRank}, \textit{Token Frequency}) showed more balanced or late-leaning distributions.

LLMs demonstrated moderate early biases. Qwen2.5 and Llama variants typically favored early positions more than the DeepSeek model. The weakest early bias (i.e., strongest late bias) varied by dataset: \textit{Token Frequency} on CNN/DailyMail, \textit{DeepSeek-Chat} on SAMSum, and \textit{TextRank} on DECODA. These findings suggest that while baseline methods rely heavily on document structure, LLMs exhibit more nuanced positional preferences that likely reflect learned content understanding rather than simple heuristics.

A comparison across datasets shows that CNN/DailyMail exhibits the strongest early-position bias, aligning with the standard inverted pyramid structure of news articles. In contrast, the two dialogue datasets display more moderate positional biases: SAMSum shows the most balanced distribution, whereas DECODA displays a comparatively stronger, yet still moderate, early bias. This difference likely reflects their distinct discourse structures: casual multi-turn conversations in SAMSum versus the goal-oriented, problem-solution sequences characteristic of customer service dialogues in DECODA.

\subsection{Model Entropy Comparison}
\label{appendix:model_entropy}

\begin{figure}[!htb]
  \includegraphics[trim=0 4cm 0 0, clip, width=\columnwidth]{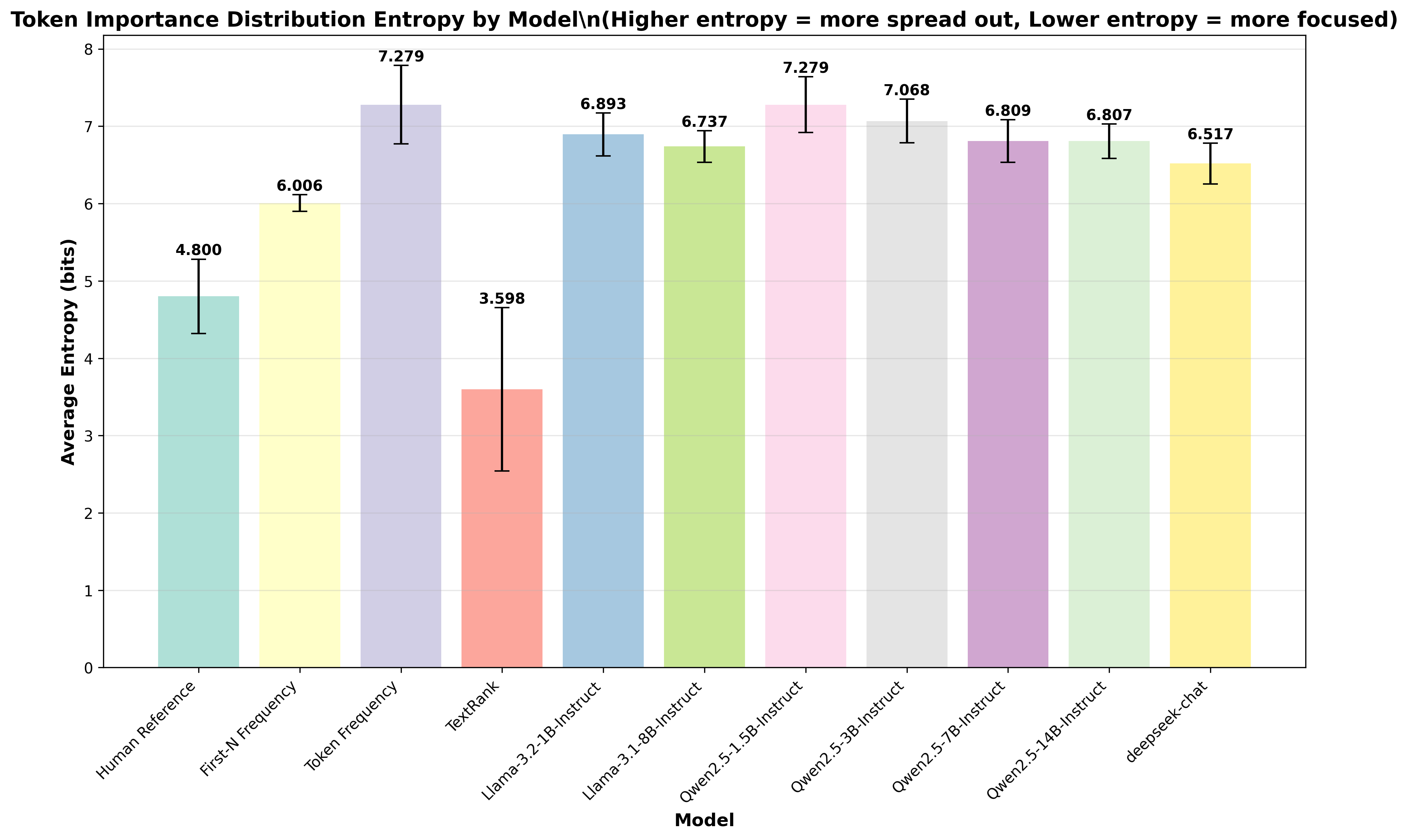} \\
  \includegraphics[trim=0 4cm 0 0.8cm, clip, width=\columnwidth]{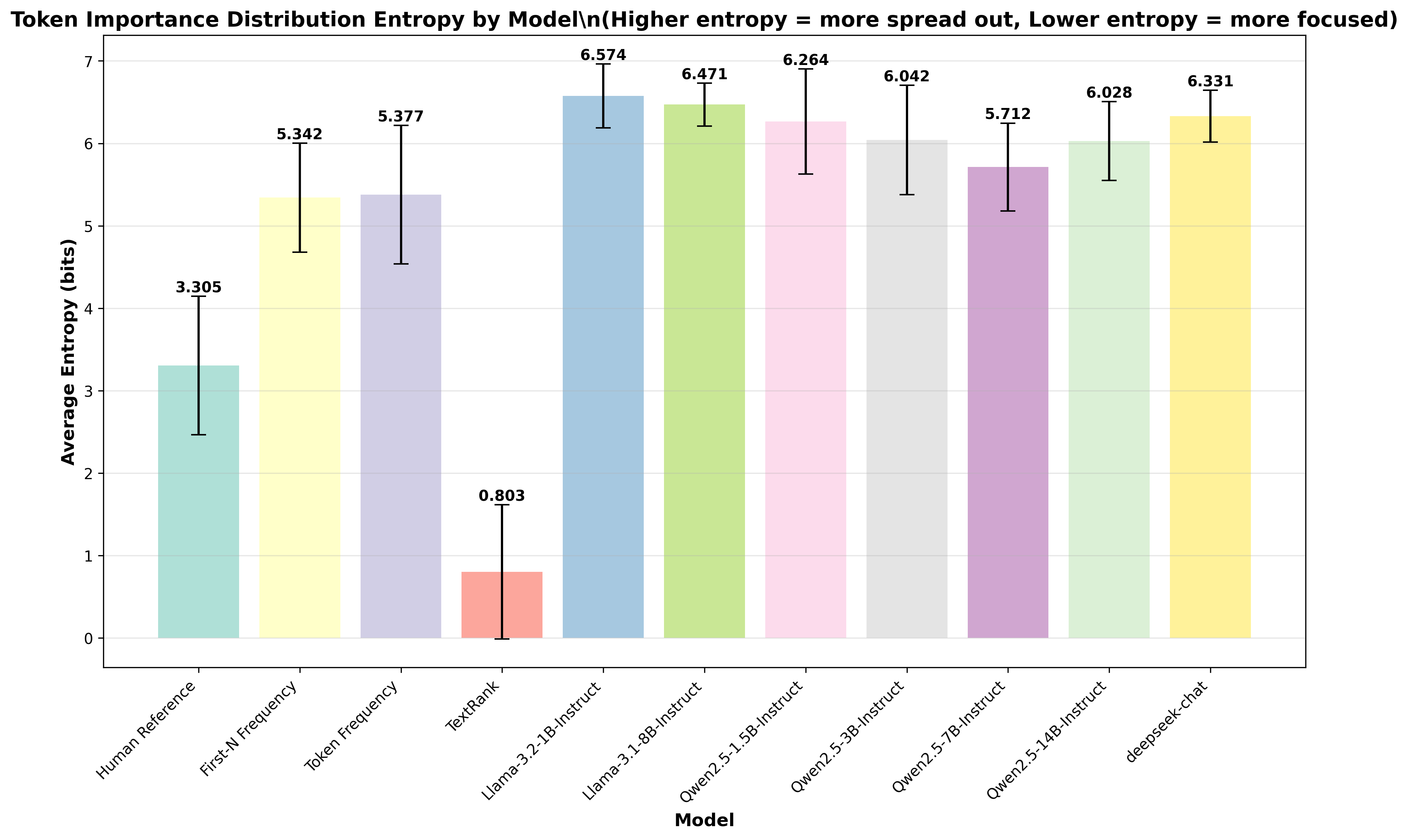} \\
  \includegraphics[trim=0 0 0 0.8cm, clip, width=\columnwidth]{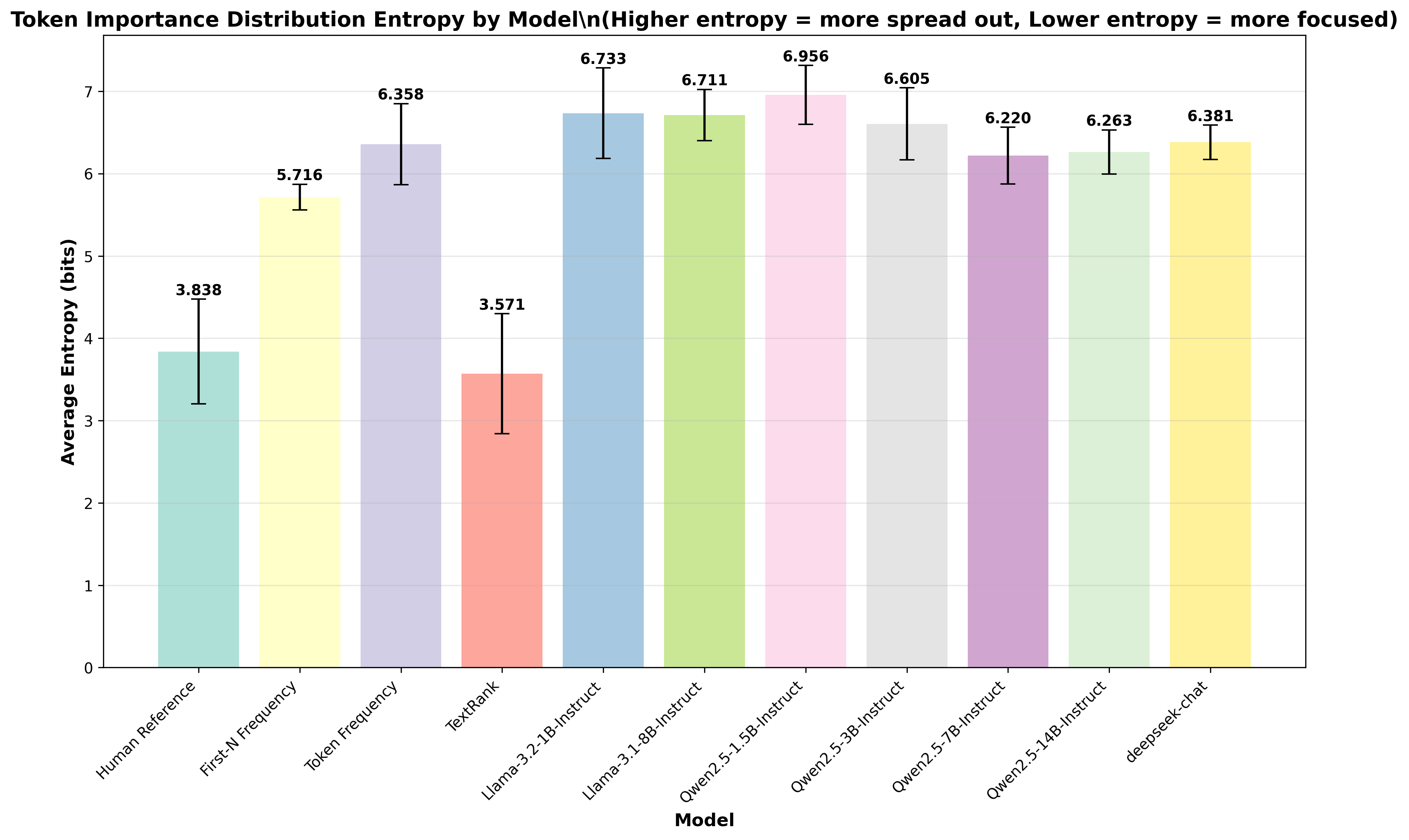}
  \caption{Entropy comparison by models on different datasets (top: CNN/DailyMail, middle: SAMSum, bottom: DECODA).}
  \label{fig:model_entropy_comparison}
\end{figure}

We quantify the concentration of model-dependent \emph{importance distribution} \( I_M(D) \) using Shannon entropy \citep{6773067}.
Figure \ref{fig:model_entropy_comparison} shows these entropy distributions across models, where lower entropy indicates importance is concentrated on fewer words, and higher entropy reflects more distributed attention.

We compute entropy as $H(I) = -\sum_i p_i \log_2(p_i)$ using the normalized importance scores $p_i = I_i / \sum_j I_j$.
We observe systematically higher entropy for models on the CNN/DailyMail dataset compared to SAMSum and DECODA, consistent with the greater length and informational density of news articles relative to conversational dialogues.

The results reveal distinct patterns across datasets. On CNN/DailyMail, DeepSeek-Chat produces the most focused distributions (lowest entropy: 6.52 $\pm$ 0.26 bits), while Qwen2.5-1.5B-Instruct yields the most dispersed (highest entropy: 7.28 $\pm$ 0.36 bits). This pattern differs on SAMSum, where Qwen2.5-7B-Instruct shows the most concentrated assignments (5.71 $\pm$ 0.53 bits) and Llama-3.2-1B-Instruct the most distributed (6.57 $\pm$ 0.39 bits). On DECODA, models maintain higher overall entropy, with Qwen2.5-1.5B-Instruct reaching 6.96 $\pm$ 0.36 bits and DeepSeek-Chat maintaining relatively selective attention at 6.38 $\pm$ 0.21 bits. This stability across languages suggests robust, language-invariant mechanisms for encoding importance.

\section{Attention–Importance Alignment}
\label{appendix:multi_head_attention}

\subsection{Heatmap Visualization}

\paragraph{NDCG@10} 

\begin{figure}[!htb]
\centering
\includegraphics[trim=0 0 0 1cm, clip, width=\columnwidth]{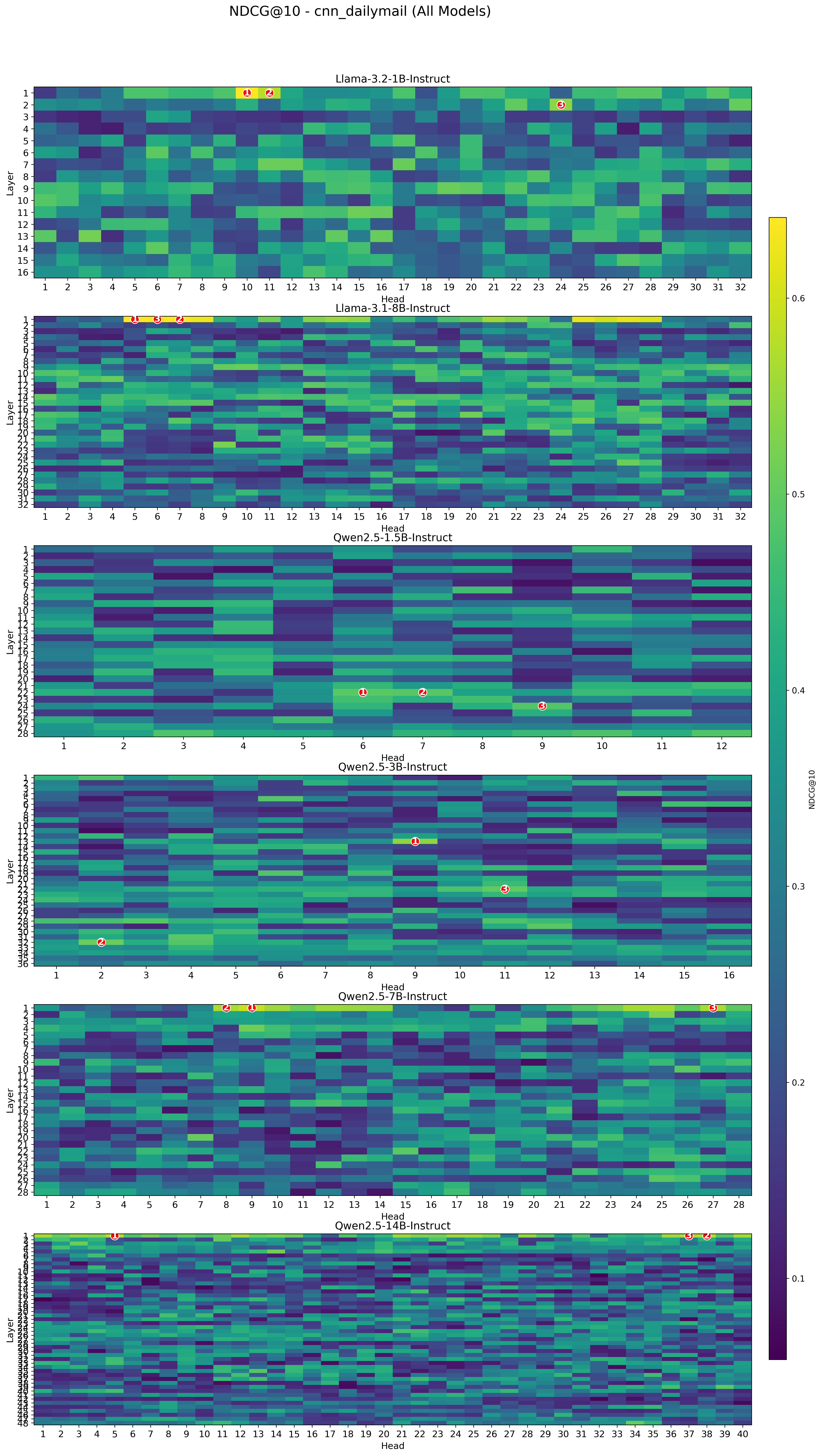}
    \caption{Average NDCG@10 per attention head on CNN/DailyMail. Red markers with ranks 1-3 indicate the top three heads per model.}
  \label{fig:attention_ndcg_cnn_dailymail}
\end{figure}

\begin{figure}[!htb]
\centering
\includegraphics[trim=0 0 0 1cm, clip, width=\columnwidth]{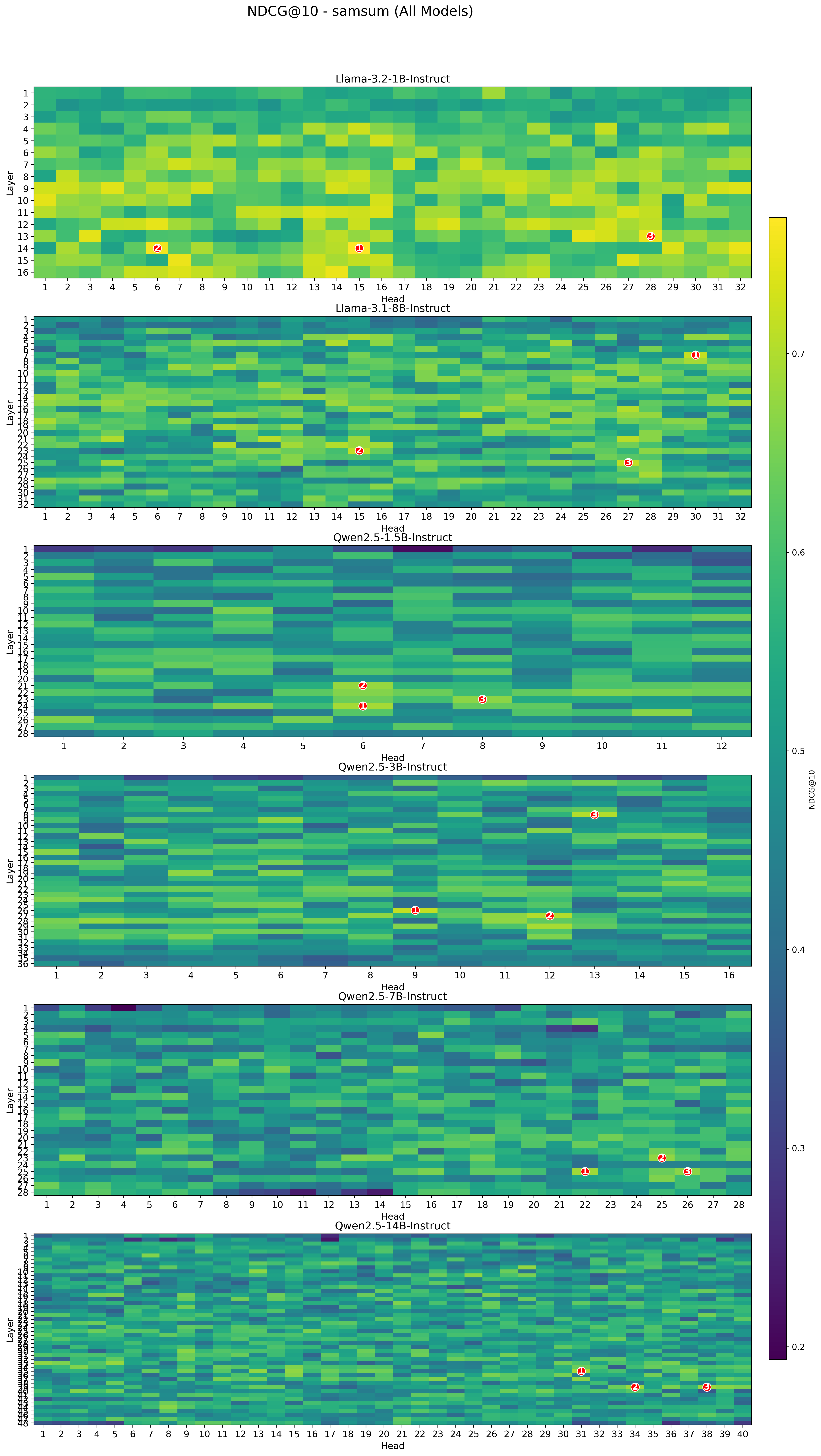}
  \caption{Average NDCG@10 per attention head on SAMSum. Red markers with ranks 1–3 indicate the top three heads per model.}
  \label{fig:attention_ndcg_samsum}
\end{figure}

\begin{figure}[!htb]
\centering
    \includegraphics[trim=0 0 0 1cm, clip, width=\columnwidth]{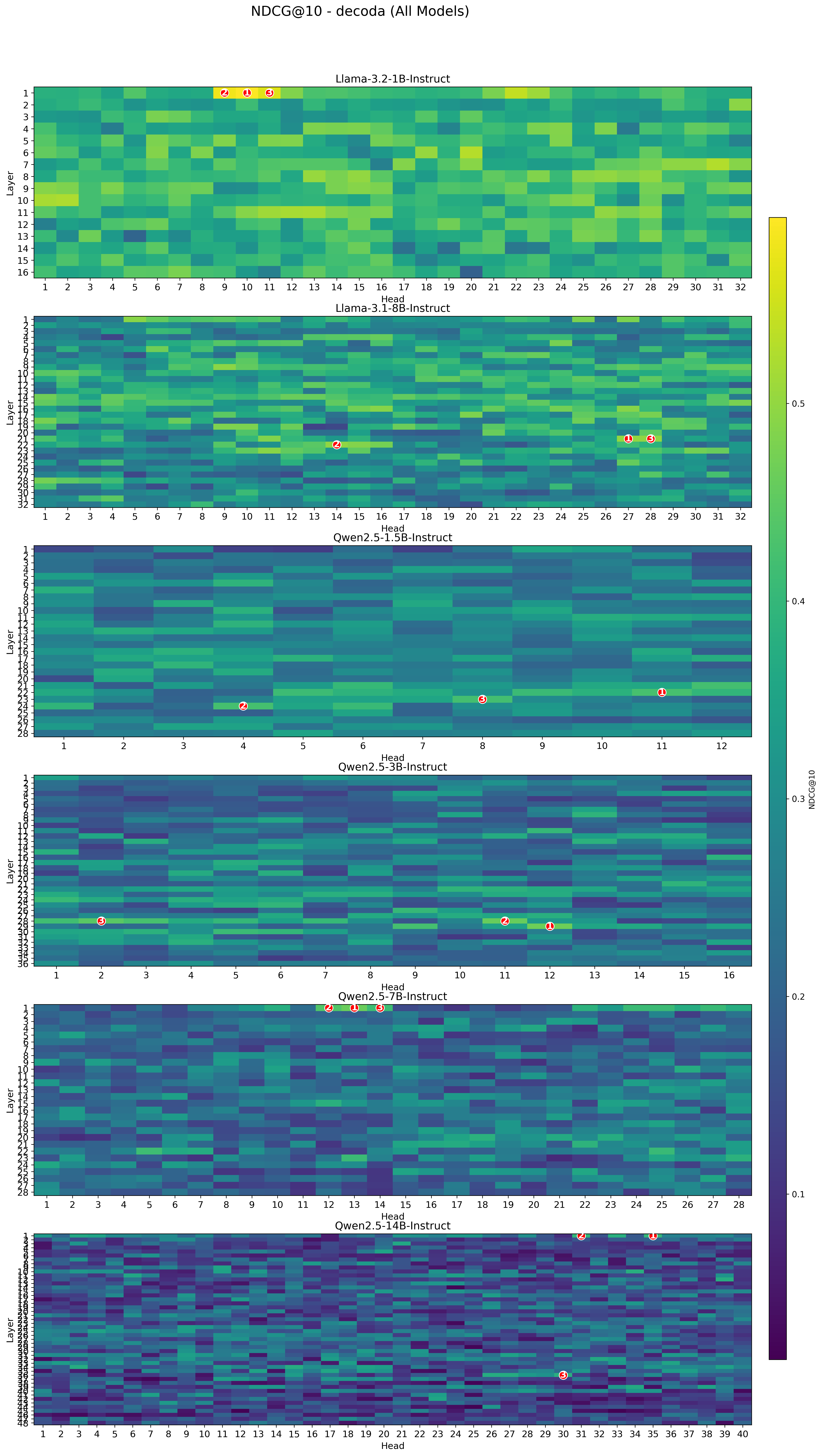}
  \caption{Average NDCG@10 per attention head on DECODA. Red markers with ranks 1–3 indicate the top three heads per model.}
  \label{fig:attention_ndcg_decoda}
\end{figure}

To compare the top-$k$ ranking consistency of attention heads across datasets, we visualize the average NDCG@10 per head in Figs. \ref{fig:attention_ndcg_cnn_dailymail}, \ref{fig:attention_ndcg_samsum}, and \ref{fig:attention_ndcg_decoda} for CNN/DailyMail, SAMSum, and DECODA, respectively.

\paragraph{Spearman rank correlation}

\begin{figure}[!htb]
\centering
\includegraphics[trim=0 0 0 1cm, clip, width=\columnwidth]{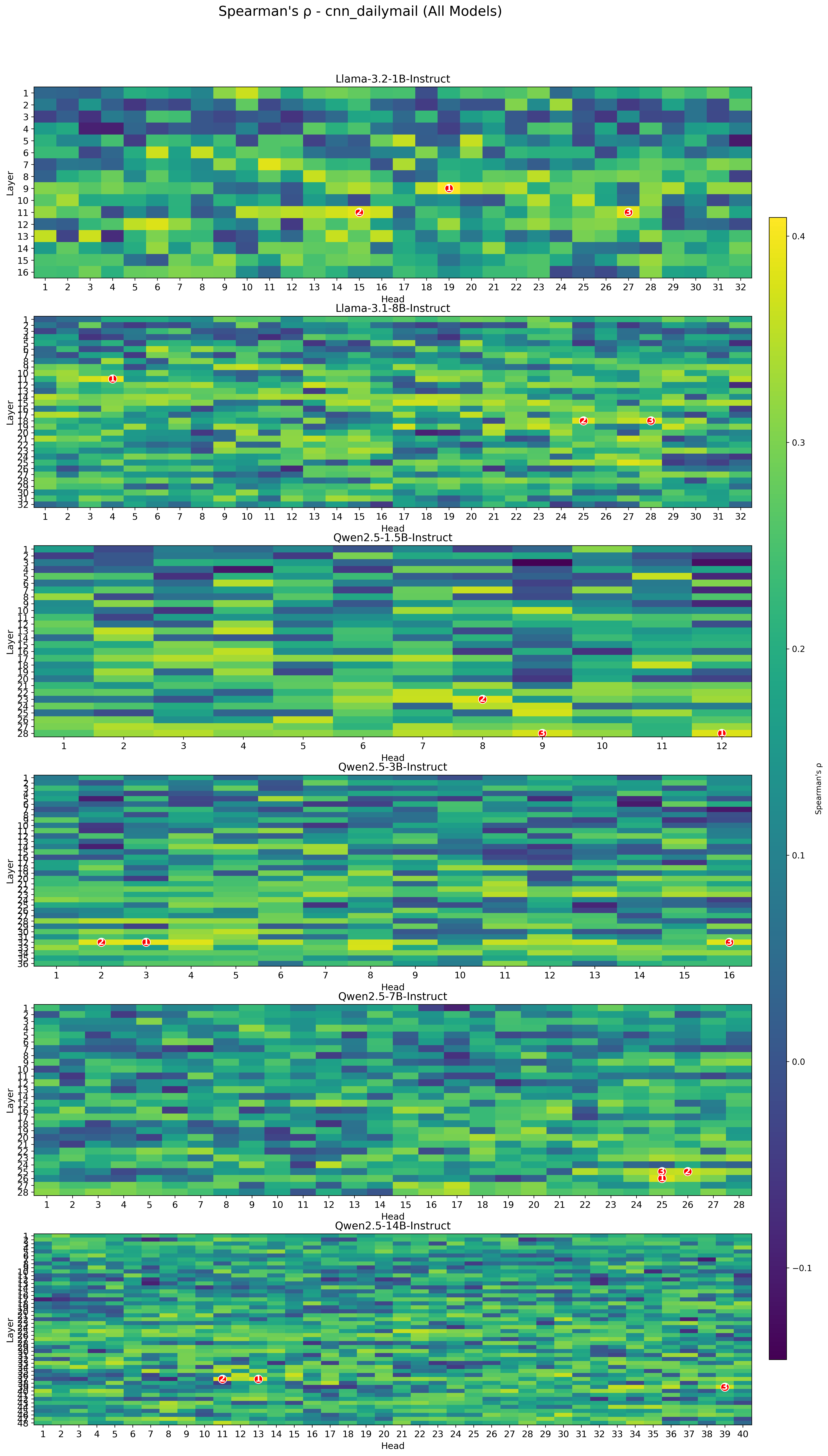}
    \caption{Average Spearman $\rho$ per attention head on CNN/DailyMail. Red markers with ranks 1–3 indicate the top three heads per model.}
  \label{fig:attention_spearman_cnn_dailymail}
\end{figure}

\begin{figure}[!htb]
\centering
\includegraphics[trim=0 0 0 1cm, clip, width=\columnwidth]{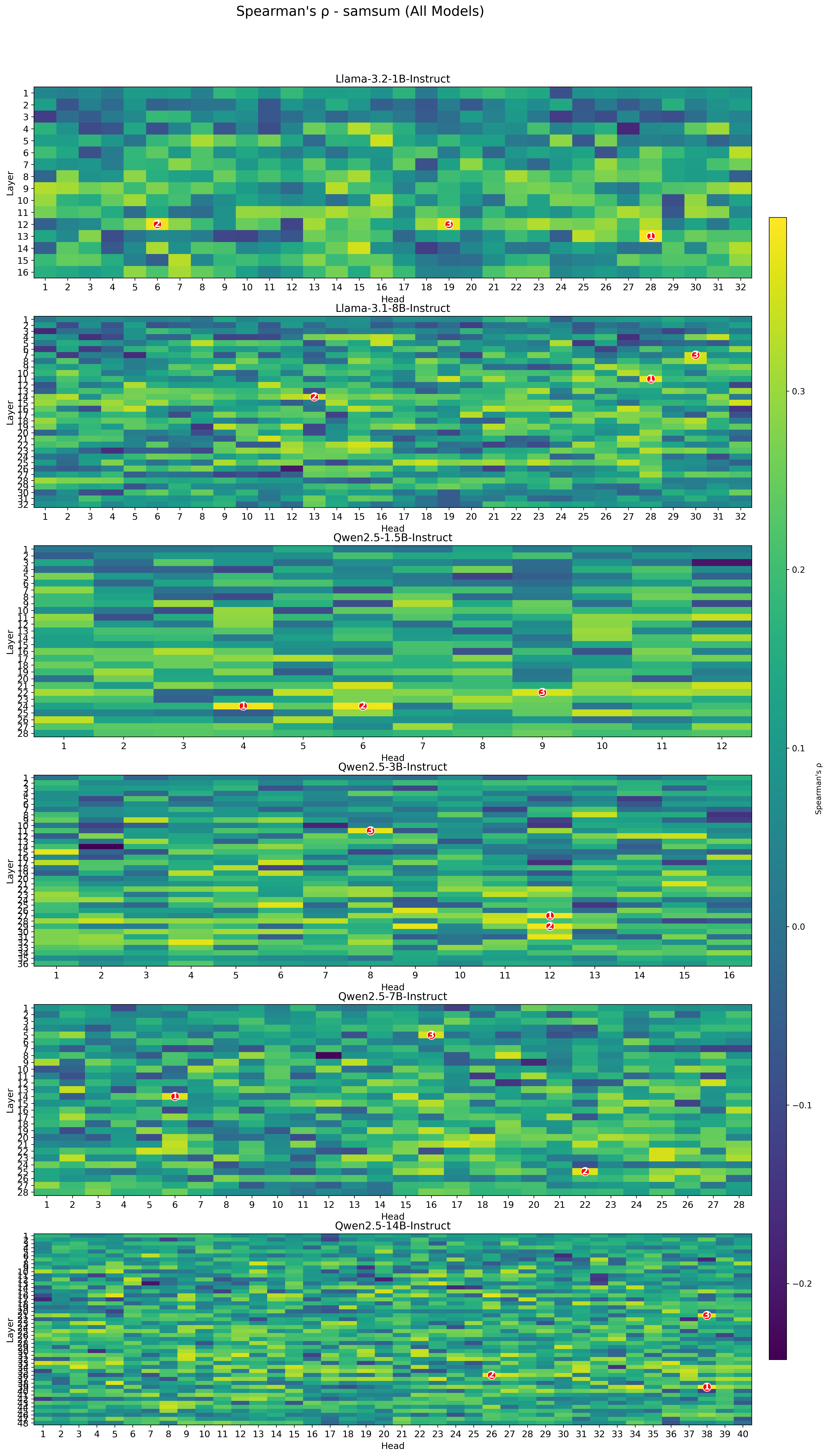}
  \caption{Average Spearman $\rho$ per attention head on SAMSum. Red markers labeled with rank 1–3 indicate the top three heads per model.}
  \label{fig:attention_spearman_samsum}
\end{figure}

\begin{figure}[!htb]
\centering
    \includegraphics[trim=0 0 0 1cm, clip, width=\columnwidth]{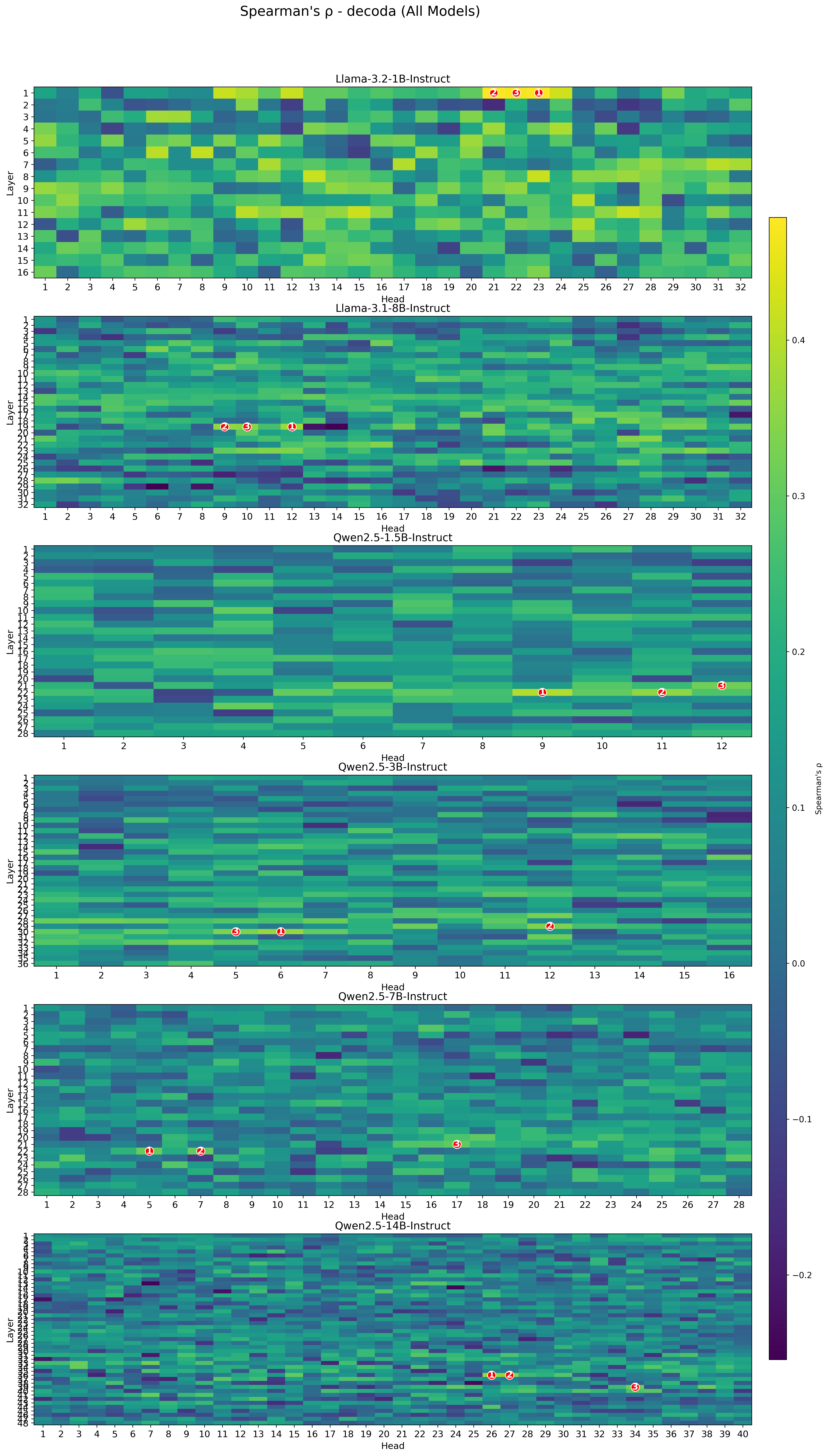}
  \caption{Average Spearman $\rho$ per attention head on DECODA. Red markers with ranks 1–3 indicate the top three heads per model.}
  \label{fig:attention_spearman_decoda}
\end{figure}

We further measure attention-importance alignment using Spearman rank correlation. Figures~\ref{fig:attention_spearman_cnn_dailymail},~\ref{fig:attention_spearman_samsum} and~\ref{fig:attention_spearman_decoda} visualize the average Spearman $\rho$ per attention head for different LLMs on the three datasets, respectively, with the top three heads per model indicated by red ranked markers. 

Our multi-head attention analysis shows that attention is not a monolithic proxy for importance. Early layers show negligible correlation, while specific heads in middle-to-late layers emerge as strong predictors (Spearman $\rho \approx 0.4$). Architecturally, Qwen2.5 models align importance in final layers, whereas Llama models align in middle layers. This suggests that \emph{information importance} is captured by specialized attention heads, not by attention as a whole.

\subsection{NDCG@10 by Layer} 

\begin{figure}[!htb]
    \centering

    \begin{subfigure}[t]{0.48\textwidth}
        \centering
        \includegraphics[width=\linewidth]{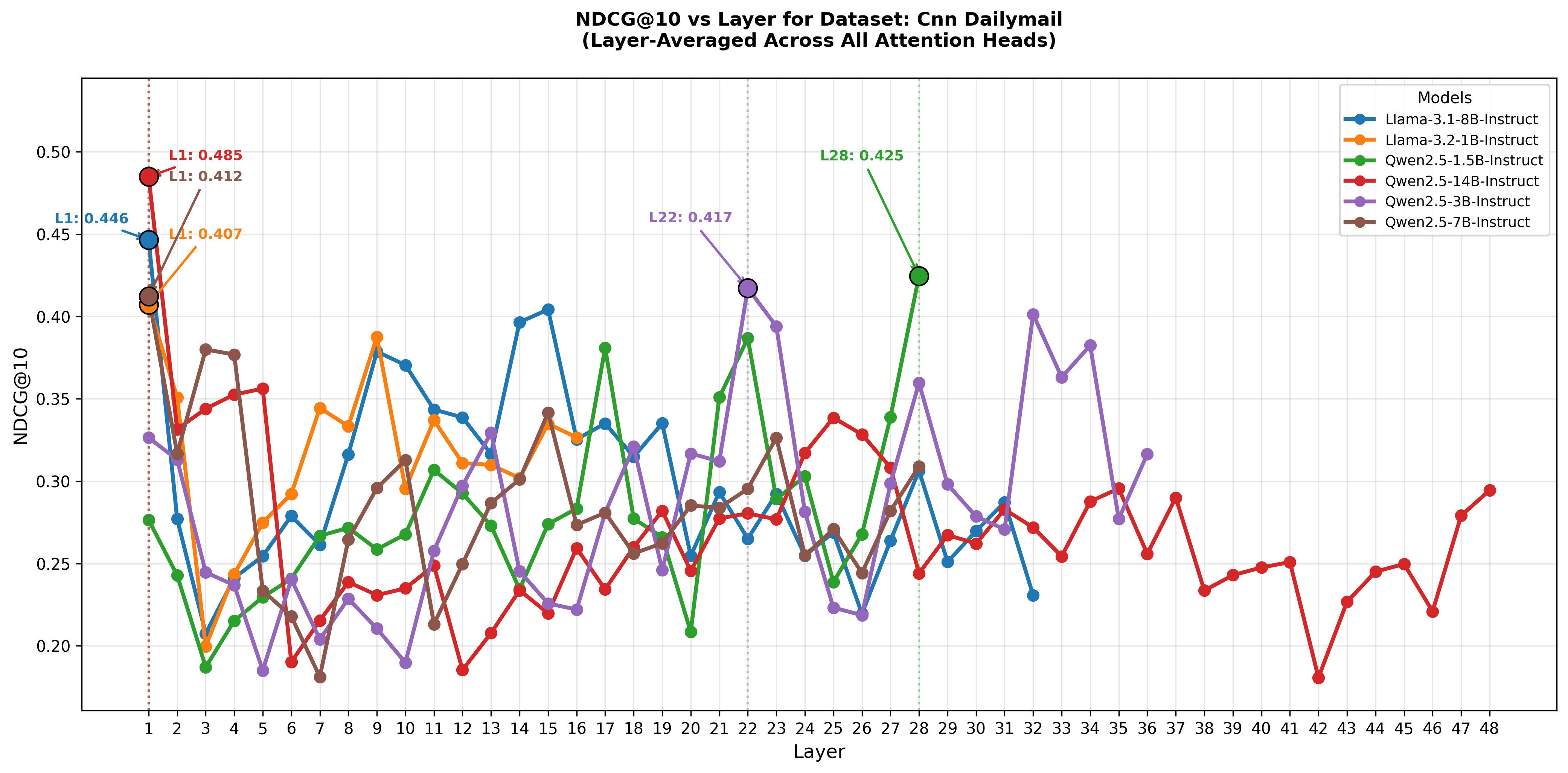}
        \caption{CNN/DailyMail}
        \label{fig:attention_ndcg_layer_cnn}
    \end{subfigure}
    \hfill 
    \begin{subfigure}[t]{0.48\textwidth}
        \centering
        \includegraphics[width=\linewidth]{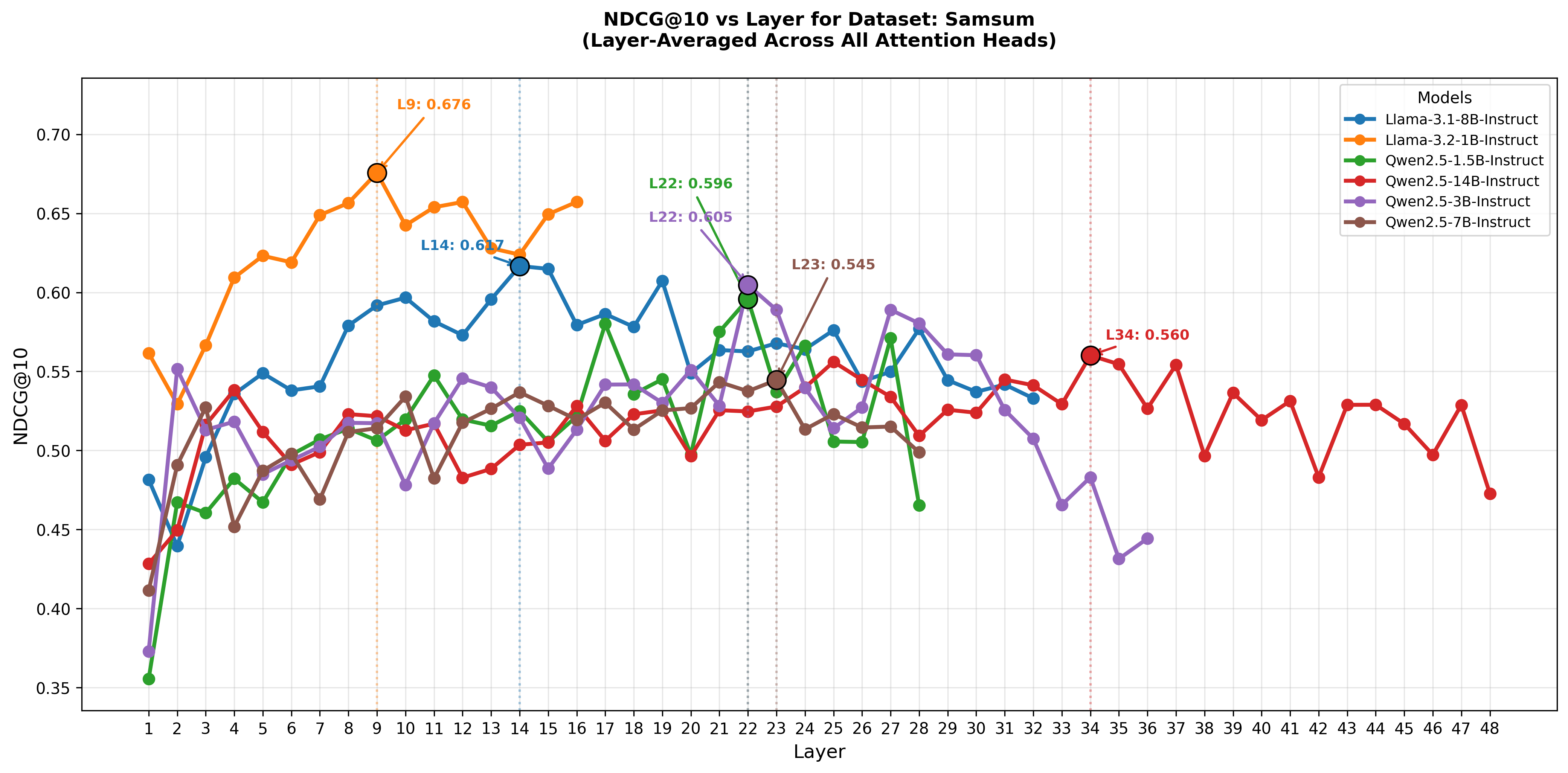}
        \caption{SAMSum}
        \label{fig:attention_ndcg_layer_samsum}
    \end{subfigure}
    \hfill
    \begin{subfigure}[t]{0.48\textwidth}
        \centering
        \includegraphics[width=\linewidth]{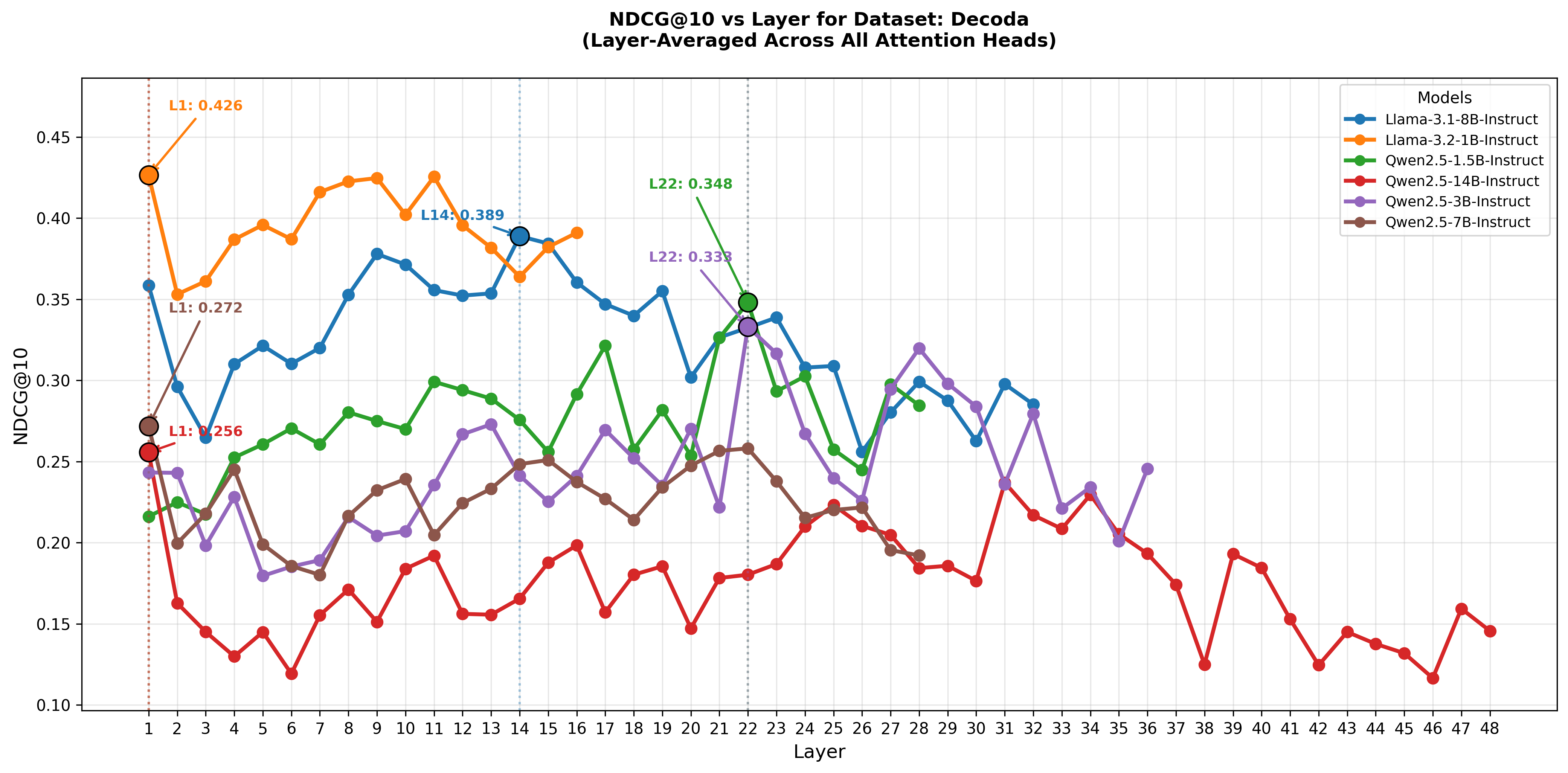}
        \caption{DECODA}
        \label{fig:attention_ndcg_layer_decoda}
    \end{subfigure}

    \caption{Evolution of average NDCG@10 scores across transformer layers for the CNN/DailyMail (a), SAMSum (b), and DECODA (c) datasets. Each point represents the mean NDCG@10 across all attention heads within a layer.}
    \label{fig:attention_ndcg_vs_layer}
\end{figure}

In Figure \ref{fig:attention_ndcg_vs_layer}, we present the evolution of average NDCG@10 scores across model layers for the CNN/DailyMail (a), SAMSum (b), and DECODA (c) datasets. Each point represents the mean NDCG@10 across all attention heads within a layer, illustrating how the alignment between attention and the model-dependent \emph{importance distribution} \( I_M(D) \) varies with network depth.

\subsection{Metrics Results for the Best Head}
\label{appendix:best_heads_metrics}

\begin{table}[!htb]
    \centering
\include{tables/best_head_metrics_results}
    \caption{Complete evaluation metrics for the best-performing attention head (Layer~13, Head~14) of Llama-3.2-1B-Instruct on the SAMSum dataset. Reported as mean $\pm$ standard deviation where applicable; values are rounded to three decimal places.}
\label{tab:complete_metrics}
\end{table}

The MDS projection in Section~\ref{subsec:attention_heads_mds_visualization} identified the best-performing attention head for Llama-3.2-1B-Instruct on SAMSum as Layer~13, Head~14 (NDCG@10 = 0.77). The results of other evaluation metrics for this head are reported in Table~\ref{tab:complete_metrics}.

This head (Layer~13, Head~14) exhibits the strongest alignment with \emph{importance distribution} \( I_M(D) \) among all 512 attention heads. The NDCG@10 score of 0.769 indicates strong consistency in the ranking of the top-10 words, while the Spearman correlation of 0.352 reflects a moderate overall rank correlation. 
In addition, high precision scores (Precision@5 = 79.4\%, Precision@10 = 68.4\%) suggest that when this head assigns high attention to words, they are likely to be important. Conversely, the low recall scores (e.g., Recall@10 = 36.3\%) indicate that it captures only a subset of all important words, signifying a specialized rather than comprehensive focus.

The standard deviations across samples quantify the consistency of these metrics. For this head, the Spearman correlation shows moderate variability (std = 0.161). The NDCG scores exhibit high consistency (std $\approx$ 0.11), indicating reliable top-$k$ ranking quality. The Rényi divergence metrics show greater variability (std $\approx$ 0.55), suggesting more sample-dependent distribution similarity. Overall, the low standard deviations, particularly for NDCG, confirm that the head's performance is stable.

\section{Probing Results for the Three Scenarios}
\label{appendix:probing_all_scenarios}

\subsection{Baselines for Comparison}
\label{appendix:probing_baselines}

TextRank baseline values for each model and dataset are reported in Tables~\ref{tab:textrank_spearman_baselines} (Spearman correlation) and~\ref{tab:textrank_ndcg_baselines} (NDCG@10). These baselines quantify the alignment between unsupervised TextRank scores and the model-dependent \emph{importance distribution} \( I_M(D) \), establishing a content-aware performance lower bound for evaluating the probes.
 
\begin{table}[!htb]
\centering
\include{tables/tbl_textrank_spearman_baselines}
\caption{TextRank Spearman correlation baselines (mean $\pm$ std) per model and dataset.}
\label{tab:textrank_spearman_baselines}
\end{table}

\begin{table}[!htb]
\centering
\include{tables/tbl_textrank_ndcg_baselines}
\caption{TextRank NDCG@10 baselines (mean $\pm$ std) per model and dataset.}
\label{tab:textrank_ndcg_baselines}
\end{table}

\subsection{Scenario 3: Article-level Probing}
\label{appendix:article_level_probing}

\begin{figure*}[htb]
\centering
  \includegraphics[trim=0 0 0 1.2cm, clip, width=.9\textwidth]{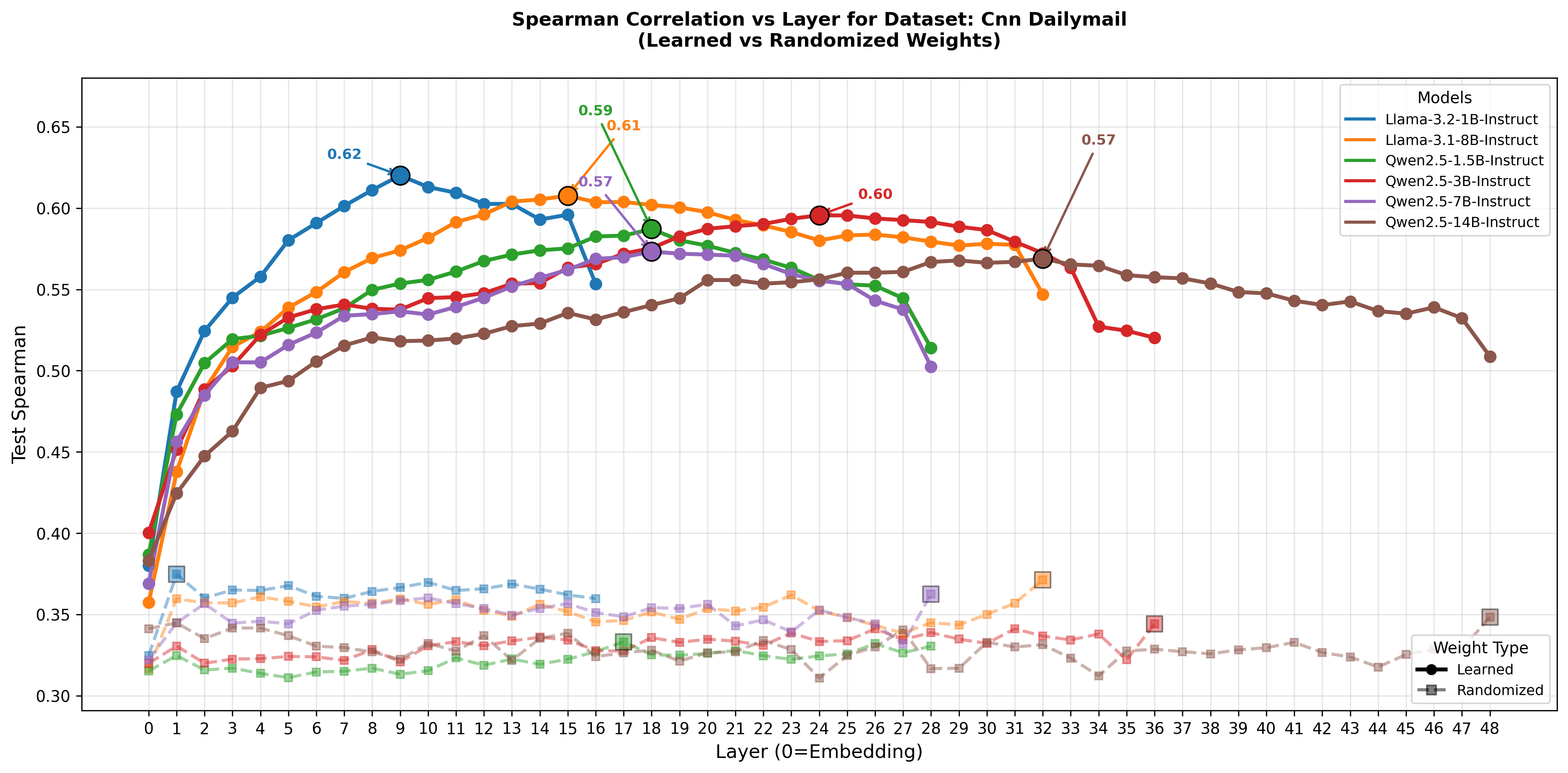}
  \includegraphics[trim=0 0 0 1.2cm, clip, width=.9\textwidth]{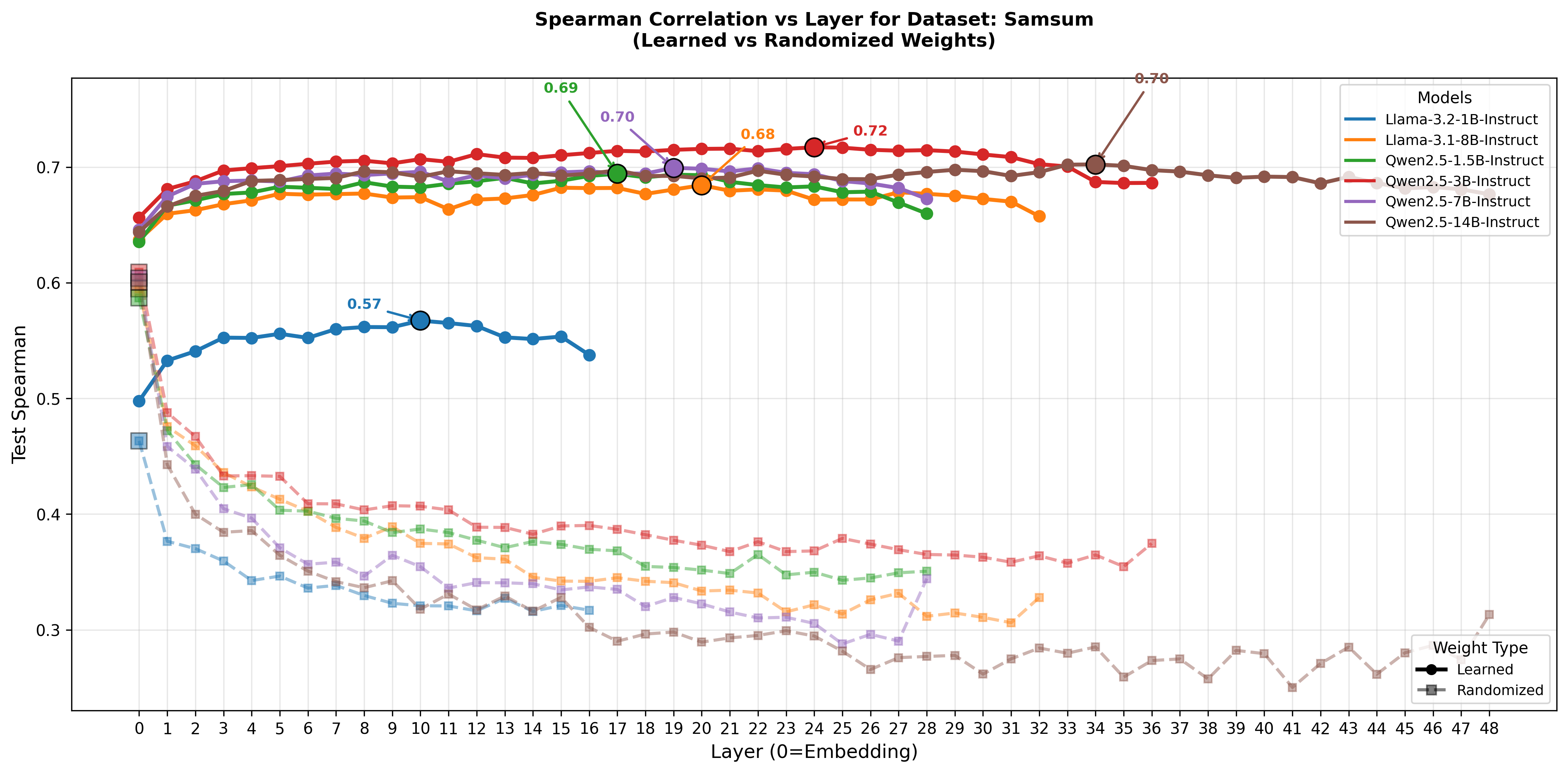}
  \includegraphics[trim=0 0 0 1.2cm, clip, width=.9\textwidth]{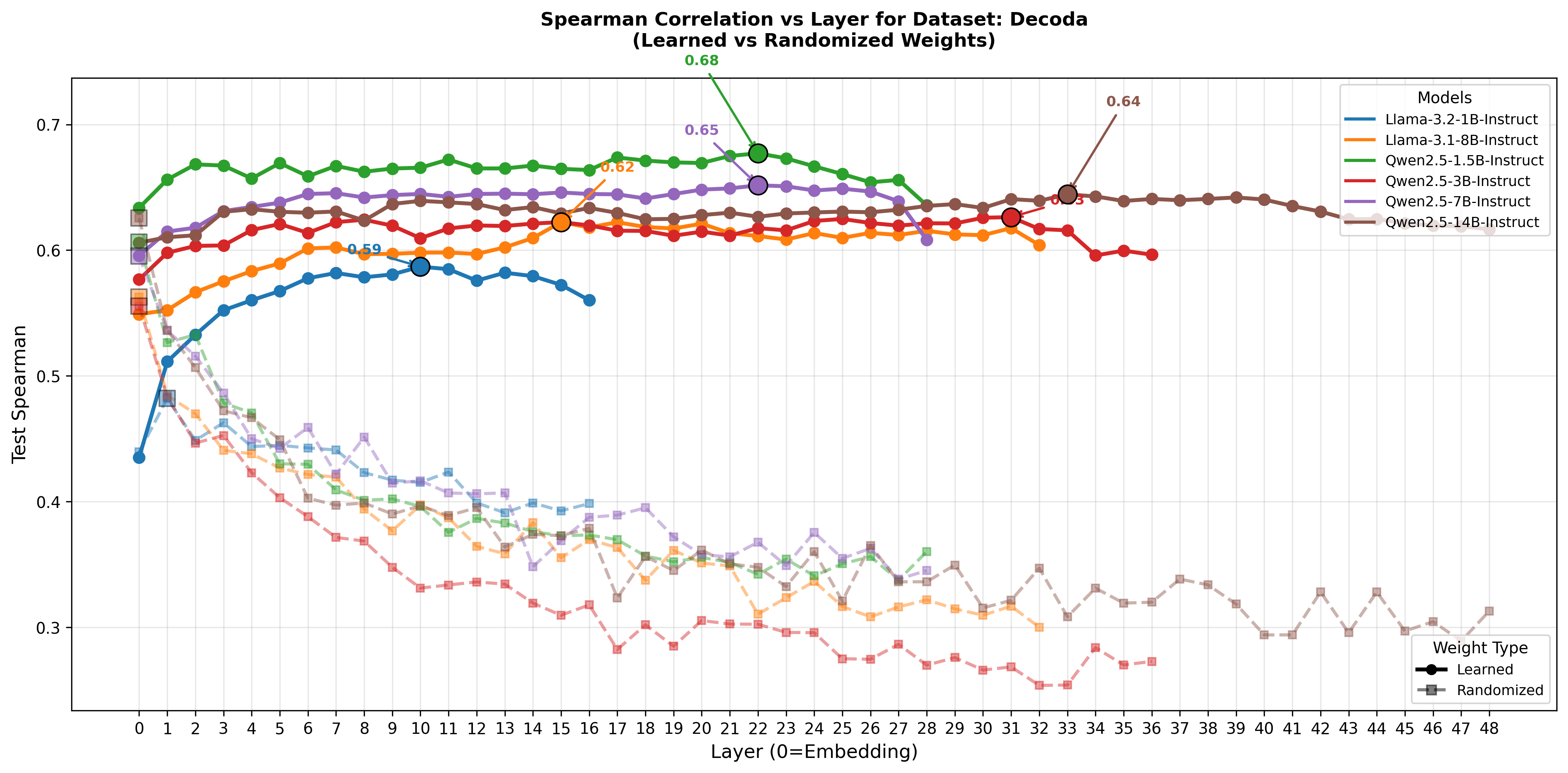}
  \caption{Spearman rank correlation ($\rho$) for article-level probing across layers on the CNN/DailyMail (top), SAMSum (middle), and DECODA (bottom) datasets. Performance is indicated by round markers, while square markers denote the \textit{Randomized Weights Baseline}. The best-performing layer for each dataset is highlighted and annotated. TextRank baseline results (approximately zero) are omitted for clarity.} 
  \label{fig:S3_article_level_spearman}
\end{figure*}

\begin{figure*}[!htb]
  \centering
  \includegraphics[trim=0 0 0 1.5cm, clip, width=.9\textwidth]{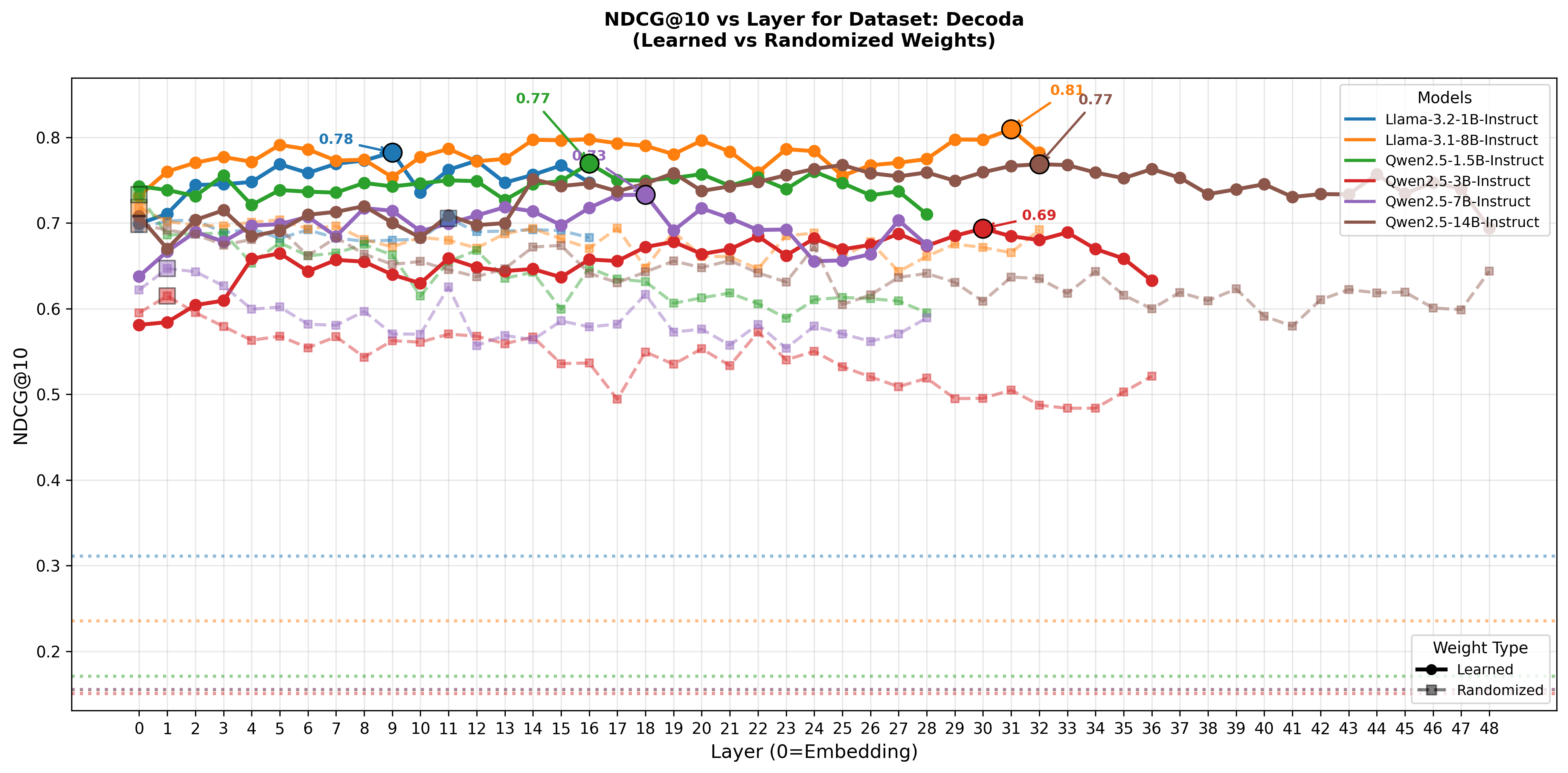}
    \caption{Article-level probing NDCG@10 across layers for DECODA. Round dots show learned model performance; square dots show the \textit{Randomized Weights Baseline}. The best-performing layers are annotated. Horizontal dashed lines show the TextRank baseline for each model.}
  \label{fig:S3_article_level_ndcg_decoda}
\end{figure*}

This section provides complementary probing analyses. In addition to the main results, we report NDCG@10 on the French summarization dataset DECODA (Figure~\ref{fig:S3_article_level_ndcg_decoda}) to examine cross-lingual robustness.
On DECODA, the performance gap between probes trained on learned and randomized weights is narrower than on the English datasets, with peak performance ($\approx$ 0.81 for Llama-3.1-8B-Instruct) occurring in middle-to-late layers. This suggests that importance ranking on DECODA may rely less on learned patterns, possibly due to more directly accessible base representations or distinct structural properties of the task.

Complementary Spearman correlation results are shown in Figure~\ref{fig:S3_article_level_spearman}.

\textbf{Overall effectiveness of information importance probing.} The figure presents layer-wise Spearman correlations between probe-predicted information importance rankings and the empirical \emph{importance distribution} \( I_M(D) \). Across all evaluated models and datasets, Spearman values range from moderate to high, indicating that hidden states encode substantial information relevant to word-level importance in summarization. Notably, even the worst-performing layers consistently outperform the TextRank baseline (see Tables~\ref{tab:textrank_spearman_baselines}), which is omitted from the figure for clarity.

\textbf{Layer-wise localization of information importance.} A consistent pattern emerges across models: peak probing performance is typically attained in the middle-to-late layers rather than at the embedding or final layers. For instance, on CNN/DailyMail, LLaMA-3.2-1B-Instruct reaches its maximum correlation at layer 9 (out of 16 layers), while larger models such as Qwen-2.5-14B-Instruct peak substantially deeper (e.g., around layer 32 out of 48 layers). A similar trend is observed on SAMSum, where most models achieve their highest correlations in the upper-middle layers. This layer-wise behavior suggests that \emph{information importance} is most explicitly represented after initial lexical processing but before the final layers, which are more specialized toward generation and output distribution modeling.

\textbf{Cross-task differences in layer sensitivity.} While the overall trend is consistent across datasets, the degree of layer-wise variation differs markedly between tasks. On CNN/DailyMail, which involves long, information-dense articles, probing performance varies substantially across layers, indicating a more pronounced redistribution of importance-related information throughout the network depth. In contrast, on SAMSum, a dialogue summarization task with shorter inputs and more localized salient content, Spearman correlations remain relatively stable across a broad range of layers. This reduced layer sensitivity suggests that importance cues in conversational data may be encoded more uniformly across representations, possibly due to lower structural complexity and weaker long-range dependencies.

Spearman correlations on the DECODA dataset remain moderate to high across models, indicating that \emph{information importance} is reliably captured by hidden representations in this French summarization setting. Similar to the trends observed on CNN/DailyMail and SAMSum, peak correlations are typically achieved in the middle-to-late layers rather than at the embedding or final layers. These results provide additional evidence that the proposed probing approach captures importance-related representations in a manner that is robust across languages.

\subsection{Scenario 1 (Layer-wise Probing)}
\label{appendix:layer_wise_probing}

\begin{figure*}[!htb]
\centering
  \includegraphics[width=.9\textwidth]{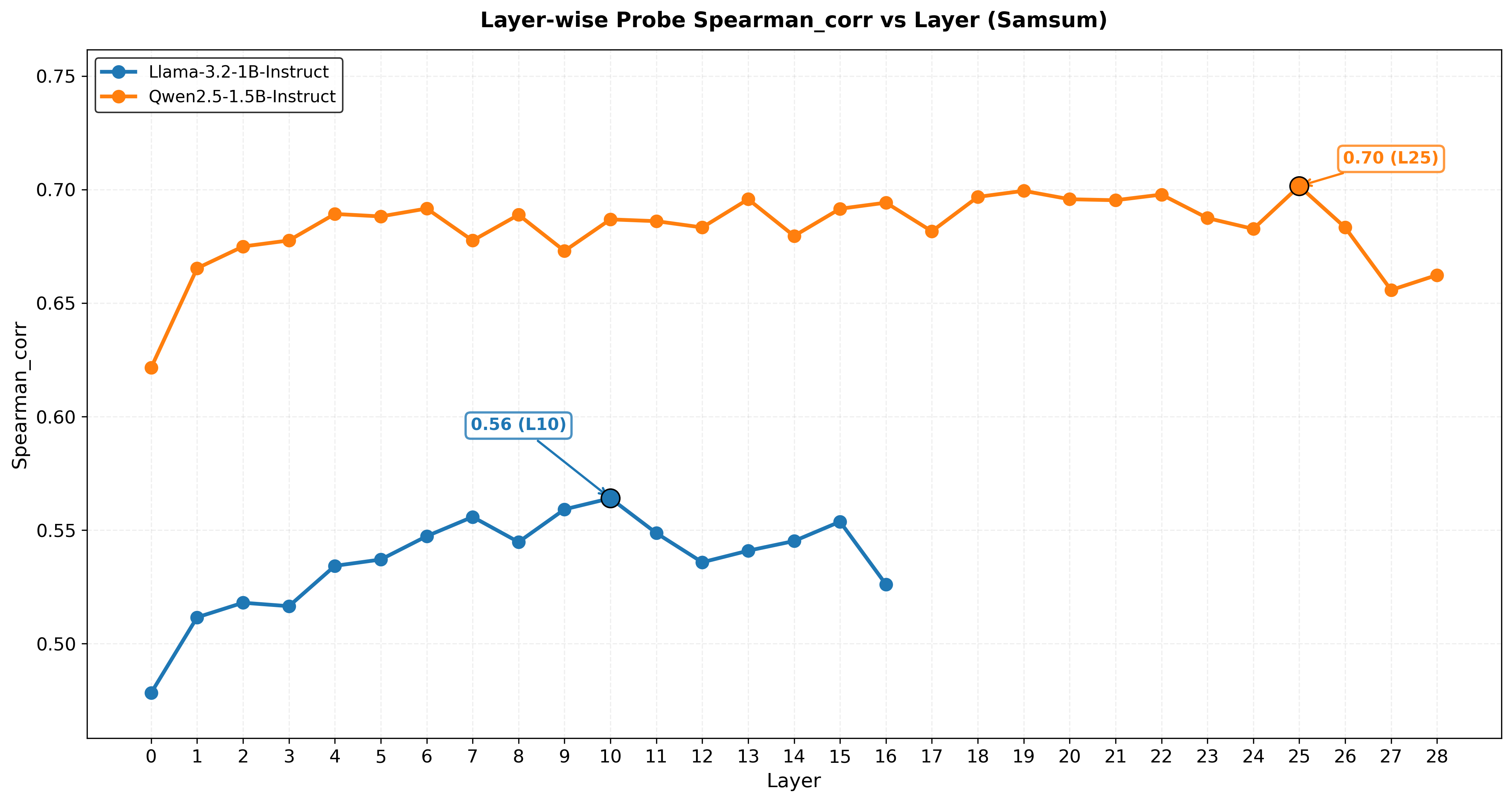}
  \caption{Layer-wise probing performance (Spearman $\rho$) across layers for Llama-3.2-1B-Instruct and Qwen2.5-1.5B-Instruct on the SAMSum dataset. Best-performing layers are annotated with their index and score. Values represent per-sample averages.} 
  \label{fig:results_layer_wise_spearman}
\end{figure*}

\begin{figure*}[!htb]
\centering
  \includegraphics[width=.9\textwidth]{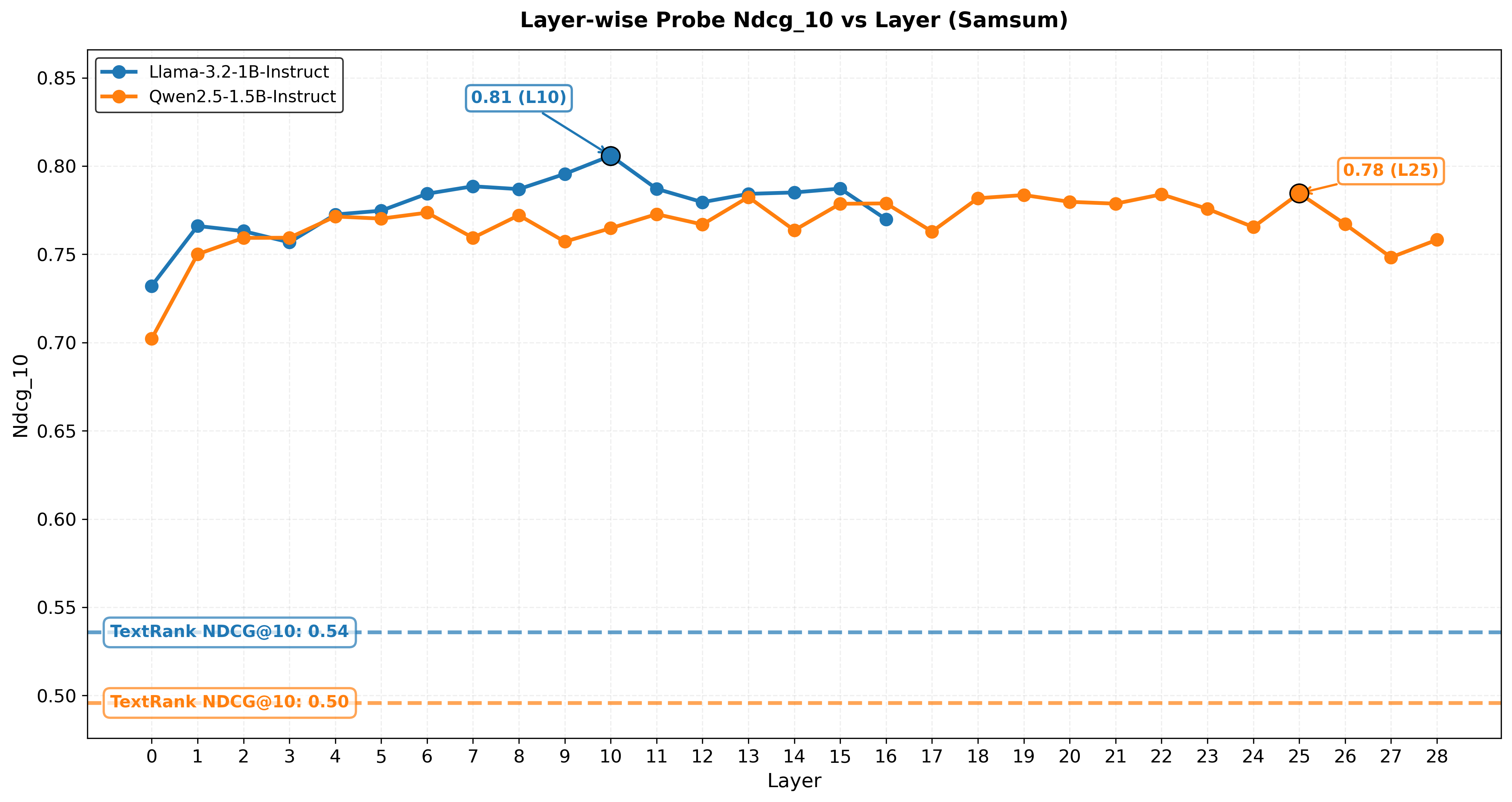}
  \caption{Layer-wise probing performance (NDCG@10) across layers for Llama-3.2-1B-Instruct and Qwen2.5-1.5B-Instruct on the SAMSum dataset. Best-performing layers are annotated with their index and score. Values represent per-sample averages.}
  \label{fig:results_layer_wise_ndcg}
\end{figure*}

For resource efficiency and because preliminary results showed consistent trends between layer-wise and all-layers probing (with middle-to-late layers generally performing best), we conducted layer-wise probing only for Llama-3.2-1B-Instruct and Qwen2.5-1.5B-Instruct on the SAMSum. The corresponding Spearman $\rho$ and NDCG@10 results are presented in Figures~\ref{fig:results_layer_wise_spearman} and \ref{fig:results_layer_wise_ndcg}.

The results reveal trends consistent with article-level probing, with best performance achieved in middle-to-late layers. Specifically, for Llama-3.2-1B-Instruct, Layer 10 performs best, while for Qwen2.5-1.5B-Instruct, Layer 25 is optimal, as indicated by both Spearman $\rho$ and NDCG@10 visualizations in Figures~\ref{fig:results_layer_wise_spearman} and~\ref{fig:results_layer_wise_ndcg}. 
Compared to all-layers probing, the best single layers yield similar but slightly lower Spearman correlations on SAMSum: peak $\rho$ values are 0.56 for Llama-3.2-1B-Instruct and 0.70 for Qwen2.5-1.5B-Instruct, compared to 0.580 and 0.715, respectively, for all-layer probes.
In terms of NDCG@10, the best single layers perform comparably: Llama-3.2-1B-Instruct achieves 0.81 versus 0.797 with all-layer probing, while Qwen2.5-1.5B-Instruct reaches 0.78 versus 0.795.

\subsection{Scenario 2 (All-layers Probing)}
\label{appendix:all_layers_probing}
In addition to the article-level and layer-wise probing experiments, we also conducted an all-layer probing by concatenating the hidden states from all layers of the model. This approach aggregates information across the network hierarchy, providing a richer representation than any single layer. As shown in Table~\ref{tab:all_layers_results}, the all-layer concatenation achieves performance comparable to the best-performing individual layer: it is slightly better in some settings, while the optimal single layer performs better in others, with overall differences remaining small.

\begin{table*}[!htb]
\centering
\include{tables/tbl_S2_probing_results}
\caption{All-layers probing results and training details across models and datasets. Performance metrics (Spearman $\rho$, NDCG@10) were computed for each sample and then averaged.}
\label{tab:all_layers_results}
\end{table*}

\subsection{Computational Experiments}

All experiments were conducted using PyTorch 2.6.0 with CUDA 12.4 on NVIDIA GPUs (RTX A6000, RTX 8000, H100). To ensure reproducibility, we fixed the random seed to 42 and enabled PyTorch's deterministic mode.

\section{Cross Dataset Transfer Capability}

\begin{figure*}[!htb]
\centering
  \includegraphics[width=.9\textwidth]{figs/sample_analysis/importance_heatmap_cnn_dailymail_id-f001ec5c4704938247d27a44948eebb37ae98d01_model-Qwen_Qwen2.5-3B-Instruct.png}
    \includegraphics[width=.9\textwidth]{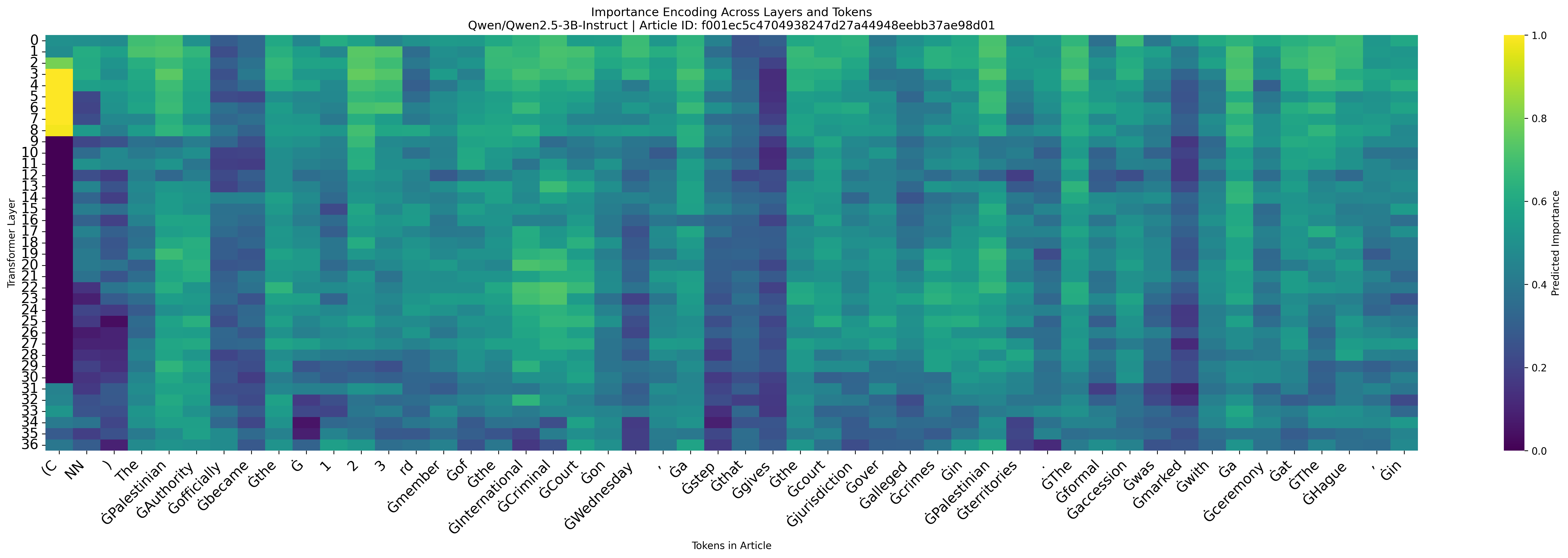}
  \caption{Layer-wise importance predictions for Qwen2.5-3B-Instruct on a CNN/DailyMail sample, comparing article-level probes trained on CNN/DailyMail (top) versus SAMSum (bottom). The heatmap displays probe outputs across all layers and the first 50 tokens.}
  \label{fig:cross_task_analysis}
\end{figure*}

We investigate the cross-dataset transfer capability of article-level probes by comparing their importance predictions on a shared CNN/DailyMail sample from Qwen2.5-3B-Instruct. As shown in Figure \ref{fig:cross_task_analysis}, the heatmap visualizes layer-wise predictions across the first 50 tokens, comparing probes trained separately on CNN/DailyMail (top) and SAMSum (bottom).

The bottom heatmap (from a probe trained on SAMSum) captures most importance patterns identified in the top heatmap (from a probe trained on CNN/DailyMail), demonstrating partial cross-dataset generalization. Both probes consistently identify key entities such as ``Palestinian'', ``Authority'', ``criminal'', and ``court'' as highly important.

However, the probe trained on SAMSum exhibits a tendency to assign higher overall importance scores compared to the CNN/DailyMail-trained probe. This suggests that the internal representation of importance may be calibrated differently when learned from the more extractive, dialogue-based data of SAMSum versus the more abstractive, long-form data of CNN/DailyMail.

\section{Licensing of Artifacts}

All datasets and models used in this work are released under the following licenses.

\begin{itemize}
    \item \textbf{CNN/DailyMail dataset} (v1.0.0): Apache-2.0 License
    \item \textbf{SAMSum dataset}\footnote{\url{https://huggingface.co/datasets/knkarthick/samsum}}: CC BY-NC-ND 4.0 (non-commercial)
    \item \textbf{DECODA dataset}: Available from the official website\footnote{\url{https://pageperso.lis-lab.fr/benoit.favre/cccs/}} upon acceptance of usage terms
    \item \textbf{Qwen models}: Apache 2.0 License
    \item \textbf{Llama models}: Llama Community License
    \item \textbf{DeepSeek-V3}\footnote{\url{https://huggingface.co/deepseek-ai/DeepSeek-V3.1-Terminus}}: MIT License
\end{itemize}

\section{AI Assistants In Research Or Writing}
This research was conducted with the assistance of AI tools for text refinement and with coding support from Copilot.

\end{document}

%% file: tables/hs_statistics.tex
\resizebox{\columnwidth}{!}{%
\begin{tabular}{lccc}
\toprule
\textbf{Model \& Dataset} & \textbf{Zero (\%)} & \textbf{Annot. (\%)} & \textbf{Words/Art.} \\
\midrule
\textbf{Llama-3.2-1B-Instruct} & & & \\
\quad CNN-DailyMail & 53.5 & \textbf{46.5} & 298.1 \\
\quad SAMSum & 22.3 & \textbf{77.7} & 82.2 \\
\midrule
\textbf{Llama-3.1-8B-Instruct} & & & \\
\quad CNN-DailyMail & 59.8 & 40.2 & 295.6 \\
\quad SAMSum & 36.8 & 63.2 & 81.8 \\
\midrule
\textbf{Qwen2.5-1.5B-Instruct} & & & \\
\quad CNN-DailyMail & 58.5 & 41.5 & 318.0 \\
\quad SAMSum & 45.8 & 54.2 & 81.4 \\
\midrule
\textbf{Qwen2.5-3B-Instruct} & & & \\
\quad CNN-DailyMail & 60.7 & 39.3 & 308.2 \\
\quad SAMSum & 44.2 & 55.8 & 77.6 \\
\midrule
\textbf{Qwen2.5-7B-Instruct} & & & \\
\quad CNN-DailyMail & 64.3 & 35.7 & 301.6 \\
\quad SAMSum & 52.2 & 47.8 & 73.3 \\
\midrule
\textbf{Qwen2.5-14B-Instruct} & & & \\
\quad CNN-DailyMail & \textbf{65.3} & 34.7 & 301.2 \\
\quad SAMSum & \textbf{49.4} & 50.6 & 75.9 \\
\bottomrule
\end{tabular}
}

%% file: tables/best_head_metrics_results.tex
\resizebox{\columnwidth}{!}{%
\begin{tabular}{lr}
\toprule
\textbf{Metric} & \textbf{Value (Mean $\pm$ Std)} \\
\midrule
\multicolumn{2}{l}{\textit{Ranking Correlation}} \\
\quad Spearman Correlation & $0.352 \pm 0.161$ \\
\midrule
\multicolumn{2}{l}{\textit{Distribution Similarity}} \\
\quad KL Score & $-4.038$ \\
\quad Power-Weighted KL Score & $2.378$ \\
\quad Rényi Divergence ($\alpha=1.5$) & $3.080 \pm 0.496$ \\
\quad Rényi Divergence ($\alpha=2.0$) & $3.496 \pm 0.543$ \\
\quad Rényi Divergence ($\alpha=3.0$) & $4.033 \pm 0.626$ \\
\midrule
\multicolumn{2}{l}{\textit{Ranking Quality}} \\
\quad NDCG@5 & $0.822 \pm 0.115$ \\
\quad NDCG@10 & $0.769 \pm 0.114$ \\
\midrule
\multicolumn{2}{l}{\textit{Information Retrieval}} \\
\quad Precision@1 & $0.986$ \\
\quad Recall@1 & $0.060$ \\
\quad Precision@3 & $0.868$ \\
\quad Recall@3 & $0.150$ \\
\quad Precision@5 & $0.794$ \\
\quad Recall@5 & $0.222$ \\
\quad Precision@10 & $0.684$ \\
\quad Recall@10 & $0.363$ \\
\bottomrule
\end{tabular}%
}

%% file: tables/tbl_textrank_spearman_baselines.tex
\resizebox{\columnwidth}{!}{%
\begin{tabular}{lccc}
\toprule
\textbf{Model} & \textbf{CNN/DailyMail} & \textbf{SAMSum} & \textbf{DECODA} \\
\midrule
\multicolumn{4}{l}{\textit{Llama Family}} \\
\quad Llama-3.2-1B-Instruct   & $-0.133 \pm 0.179$ & $0.130 \pm 0.112$ & $-0.057 \pm 0.146$ \\
\quad Llama-3.1-8B-Instruct   & $-0.162 \pm 0.194$ & $0.102 \pm 0.124$ & $-0.165 \pm 0.131$ \\
\midrule
\multicolumn{4}{l}{\textit{Qwen Family}} \\
\quad Qwen2.5-1.5B-Instruct   & $-0.145 \pm 0.175$ & $0.092 \pm 0.137$ & $-0.261 \pm 0.094$ \\
\quad Qwen2.5-3B-Instruct     & $-0.158 \pm 0.184$ & $0.106 \pm 0.139$ & $-0.290 \pm 0.128$ \\
\quad Qwen2.5-7B-Instruct     & $-0.188 \pm 0.182$ & $0.099 \pm 0.158$ & $-0.352 \pm 0.120$ \\
\quad Qwen2.5-14B-Instruct    & $-0.194 \pm 0.192$ & $0.095 \pm 0.146$ & $-0.351 \pm 0.107$ \\
\bottomrule
\end{tabular}%
}

%% file: tables/tbl_textrank_ndcg_baselines.tex
\resizebox{\columnwidth}{!}{%
\begin{tabular}{lccc}
\toprule
\textbf{Model} & \textbf{CNN/DailyMail} & \textbf{SAMSum} & \textbf{DECODA} \\
\midrule
\multicolumn{4}{l}{\textit{Llama Family}} \\
\quad Llama-3.2-1B-Instruct   & $0.446 \pm 0.156$ & $0.536 \pm 0.132$ & $0.311 \pm 0.099$ \\
\quad Llama-3.1-8B-Instruct   & $0.443 \pm 0.162$ & $0.531 \pm 0.118$ & $0.236 \pm 0.091$ \\
\midrule
\multicolumn{4}{l}{\textit{Qwen Family}} \\
\quad Qwen2.5-1.5B-Instruct   & $0.424 \pm 0.166$ & $0.496 \pm 0.121$ & $0.171 \pm 0.074$ \\
\quad Qwen2.5-3B-Instruct     & $0.428 \pm 0.163$ & $0.553 \pm 0.121$ & $0.151 \pm 0.073$ \\
\quad Qwen2.5-7B-Instruct     & $0.427 \pm 0.163$ & $0.545 \pm 0.126$ & $0.155 \pm 0.069$ \\
\quad Qwen2.5-14B-Instruct    & $0.420 \pm 0.161$ & $0.550 \pm 0.126$ & $0.155 \pm 0.076$ \\
\bottomrule
\end{tabular}%
}

%% file: tables/tbl_S2_probing_results.tex
\resizebox{\textwidth}{!}{%
\begin{tabular}{lcccccccc}
\toprule
\textbf{Model} & \textbf{Dataset} & \textbf{Spearman} & \textbf{NDCG@10} & \textbf{Input Dim} & \textbf{Dataset Size} & \textbf{Epochs} & \textbf{Early Stop} & \textbf{Time (min)} \\
\midrule
\rowcolor{gray!15}  
Llama-3.2-1B-Instruct   & CNN/DailyMail  & 0.664 & \textbf{0.835} &  34,816 & 89,421 & 5 & Yes & 191.6 \\
               & SAMSum         & 0.580 & 0.797 &  34,816 & 67,333 & 5 & Yes &  48.7 \\
\midrule
\rowcolor{gray!15}  
Llama-3.1-8B-Instruct   & CNN/DailyMail  & \textbf{0.668} & 0.828 & 135,168 & 88,674 & 5 & Yes & 255.8 \\
               & SAMSum         & 0.733 & 0.821 & 135,168 & 66,962 & 6 & Yes &  67.4 \\
\midrule
\rowcolor{gray!15}  
Qwen2.5-1.5B-Instruct      & CNN/DailyMail  & 0.636 & 0.798 &  44,544 & 95,396 & 5 & Yes & 232.3 \\
               & SAMSum         & 0.715 & 0.795 &  44,544 & 66,641 & 6 & Yes &  62.0 \\
\midrule
\rowcolor{gray!15}  
Qwen2.5-3B-Instruct        & CNN/DailyMail  & 0.643 & 0.796 &  75,776 & 92,475 & 5 & Yes & 232.9 \\
               & SAMSum         & 0.736 & 0.819 &  75,776 & 63,574 & 5 & Yes &  54.8 \\
\midrule
\rowcolor{gray!15}  
Qwen2.5-7B-Instruct        & CNN/DailyMail  & 0.648 & 0.818 & 103,936 & 90,476 & 5 & Yes & 236.9 \\
               & SAMSum         & \textbf{0.739} & \textbf{0.827} & 103,936 & 60,045 & 5 & Yes &  54.3 \\
\midrule
\rowcolor{gray!15}  
Qwen2.5-14B-Instruct       & CNN/DailyMail  & 0.639 & 0.798 & 250,880 & 90,351 & 5 & Yes & 153.2 \\
               & SAMSum         &  0.735 & 0.822 & 250,880 & 62,200 & 5 & Yes &  73.9 \\
\bottomrule
\end{tabular}
}